\documentclass[10pt,journal,compsoc]{IEEEtran}
\usepackage[nocompress]{cite}
\usepackage[pdftex]{graphicx}
\usepackage{amsmath}
\usepackage{amsthm}
\usepackage{amsfonts}
\usepackage{algorithm}
\usepackage{algorithmic}
\usepackage{hyperref}
\hypersetup{
     colorlinks   = true,
}

\newtheorem{theorem}{Theorem}

\newtheorem{proposition}{Proposition}
\newtheorem{definition}{Definition}

\usepackage[caption=false,font=footnotesize,labelfont=sf,textfont=sf]{subfig}

\begin{document}
\title{Label Efficient Regularization and Propagation for 
Graph Node Classification}

\author{Tian~Xie,~\IEEEmembership{Graduate~Student~Member,~IEEE}, 
        Rajgopal~Kannan, 
        and~C.-C.~Jay~Kuo,~\IEEEmembership{Fellow,~IEEE}%
\IEEEcompsocitemizethanks{
\IEEEcompsocthanksitem Tian~Xie and C.-C.~Jay~Kuo are with the 
Ming Hsieh Department of Electrical and Computer
Engineering, University of Southern California, Los Angeles, CA 90089,
USA, e-mails: xiet@usc.edu (Tian Xie) and cckuo@ee.usc.edu (C.-C. Jay Kuo).  
\IEEEcompsocthanksitem Rajgopal Kannan is with U.S. Army Research
Lab, Los Angeles, CA 90094 USA, e-mail: rajgopal.kannan.civ@mail.mil.
}%

% \thanks{Manuscript received April 19, 2005; revised August 26, 2015.}
}

\IEEEtitleabstractindextext{%
\begin{abstract}

An enhanced label propagation (LP) method called GraphHop was proposed
recently. It outperforms graph convolutional networks (GCNs) in the
semi-supervised node classification task on various networks.  Although
the performance of GraphHop was explained intuitively with joint node
attribute and label signal smoothening, its rigorous mathematical treatment
is lacking. In this paper, we propose a label efficient regularization
and propagation (LERP) framework for graph node classification, and
present an alternate optimization procedure for its solution.
Furthermore, we show that GraphHop only offers an approximate solution
to this framework and has two drawbacks. First, it includes all nodes in
the classifier training without taking the reliability of pseudo-labeled
nodes into account in the label update step.  Second, it provides a
rough approximation to the optimum of a subproblem in the label
aggregation step. Based on the LERP framework, we propose a new method,
named the LERP method, to solve these two shortcomings. LERP determines
reliable pseudo-labels adaptively during the alternate optimization and
provides a better approximation to the optimum with computational
efficiency.  Theoretical convergence of LERP is guaranteed.  Extensive
experiments are conducted to demonstrate the effectiveness and
efficiency of LERP. That is, LERP outperforms all benchmarking methods,
including GraphHop, consistently on five test datasets and an object
recognition task at extremely low label rates (i.e., 1, 2, 4, 8, 16, and
20 labeled samples per class). 

\end{abstract}

}

\maketitle

\IEEEdisplaynontitleabstractindextext
\IEEEpeerreviewmaketitle

\IEEEraisesectionheading{\section{Introduction}\label{sec:introduction}}
\IEEEPARstart{S}{emi-supervised} learning that exploits both labeled and
unlabeled data in learning tasks is highly in demand in real-world
applications due to the expensive cost of data labeling and the
availability of a large number of unlabeled samples. For graph problems
where very few labels are available, the geometric or manifold structure
of unlabeled data can be leveraged to improve the performance of
classification, regression, and clustering algorithms.  Graph
convolutional networks (GCNs) have been accepted by many as the
\textit{de facto} tool in addressing graph semi-supervised learning
problems \cite{kipf2016semi, hamilton2017inductive, song2022graph,
xie2022bgnn}.  Simply speaking, based on the input feature space, each
layer of GCNs applies transformation and propagation through the graph
and nonlinear activation to node embeddings. The GCN parameters are
learned via label supervision through backpropagation
\cite{rumelhart1986learning}.  The joint attributes encoding and
propagation as smoothening regularization enables GCNs to produce
a prominent performance on various real-world networks. 

There are, however, remaining challenges in GCN-based semi-supervised
learning.  First, GCNs require a sufficient number of labeled samples
for training. Instead of deriving the label embeddings from the graph
regularization as traditional methods do \cite{zhu2003semi,
zhou2003learning}, GCNs need to learn a series of projections from the
input feature space to the label space. These embedding transformations
largely depend on ample labeled samples for supervision. Besides,
nonlinear activation prevents GCNs from provable convergence. The lack of
sufficient labeled samples may make training unstable (or even
divergent). To improve it, one may integrate other semi-supervised
learning techniques (e.g., self- and co-training \cite{li2018deeper})
with GCN training or enhance the filter power to strengthen the
regularization effect \cite{li2019label}. They are nevertheless
restricted by internal deficiencies of GCNs.  Second, GCNs usually consist
of two convolutional layers and, as a result, the information from a
small neighborhood for each node is exploited \cite{kipf2016semi,
velickovic2018graph, hamilton2017inductive, abu2019mixhop}. The learning
ability is handicapped by ignoring correlations from nodes of longer
distances. Although increasing the number of layers could be a remedy,
this often leads to an oversmoothening problem and results in inseparable
node representations \cite{li2018deeper, zeng2021decoupling}.
Furthermore, a deeper model needs to train more parameters which
requires even more labeled samples. 

Inspired by the traditional PageRank \cite{page1999pagerank} and label
propagation (LP) algorithms \cite{zhu2003semi, zhou2003learning}, some
work discards feature transformations at every layer but maintains a few
with additional embedding propagation \cite{klicpera2018predict,
wu2019simplifying, huang2020combining, zhu2020simple}. This augments
regularization strength by capturing longer distance relationships and
reduces the number of training parameters for label efficiency.
Nonetheless, labels still serve as the guidance to model parameters
training rather than direct supervision in the label embedding space. As
a result, the methods still suffer the unstable training
problem and lack of supervision at very low label rates.  Along with this
research idea, the C\&S method \cite{huang2020combining} applied the
optimization procedure to the label space directly with the same
propagation as given in \cite{zhou2003learning}. Yet, it was originally
designed for supervised learning, and its performance deteriorates
significantly if the number of labeled samples decreases. Also, a simple
propagation strategy in Laplacian learning may suffer from the
degeneracy issue; namely, the solution is localized as spikes near the
labeled samples and almost constant far from them
\cite{mai2021consistent, shi2017weighted, calder2020poisson}. This is
caused by a naive propagation procedure where the information cannot be
carried over longer distances. 

An enhancement of traditional label propagation, called GraphHop, was
recently proposed in \cite{xie2022graphhop}.  GraphHop achieves
state-of-the-art performance as compared with GCN-based methods
\cite{kipf2016semi, li2018deeper, velickovic2018graph} and other
classical propagation-based algorithms \cite{zhou2003learning,
wang2007label, nie2010general}. It performs particularly well at
extremely low label rates.  Unlike GCNs that integrate node attributes
and smoothening regularization into one end-to-end training system,
GraphHop adopts a two-stage learning mechanism. Its initialization stage
extracts smoothened node attributes and exploit them to predict label
distributions of each node through logistic regression (LR) classifiers.
The predicted label embeddings can be viewed as signals on graphs, and
additional smoothening steps are applied in the subsequent iteration
stage consisting of two steps: 1) label aggregation and 2) label update.
In the first step, label embeddings of local neighbors are aggregated,
corresponding to low-pass filtering in the label space. To address
ineffective propagation in extremely low label rates, LR classifiers are
used in the second step for parameter sharing. These two steps are
conducted sequentially in each iteration for label signal smoothening.
Although the superior performance of GraphHop was explained from the
signal processing viewpoint (e.g., low-pass filtering on both attribute
and label signals) in \cite{xie2022graphhop}, its mathematical treatment
was not rigorous.

In this work, we first analyze the superior performance of GraphHop from
a variational viewpoint, which is mathematically rigorous and
transparent. To achieve this, we present a certain optimization problem
with probability constraints under the regularization framework. In
addition, we propose an alternate optimization strategy for solutions,
leading to an optimization of two non-convex subproblems. In the
meantime, we show the connection that the two steps (i.e., label
aggregation and label update) in the iteration stage of GraphHop can be
interpreted as an optimization process for these two
subproblems under convex transformation. 

Based on this variational interpretation, we point out two drawbacks of
GraphHop. First, the LR classifiers adopt all pseudo-labels into
training in the label update step, which usually suffers from the wrong
predictions. Second, only one aggregation of label embeddings is
performed in the label aggregation step, which is a rough approximation
to the optimum. We propose a new method called ``label efficient
regularization and propagation (LERP)" to address these two
shortcomings.  First, LERP determines reliable pseudo-labeled nodes
adaptively based on the distance to the labeled nodes in each alternate
round.  Second, label embeddings in LERP are propagated in several
iterations for a better and efficient approximation to the solution. 

The main contributions of this work are summarized below. 
\begin{enumerate}
\item A label efficient regularization and propagation framework with
its alternate optimization solution are proposed. Additionally, we show
that GraphHop can be analogized to this alternate procedure with convex
transformation to the subproblems. 
\item Based on the theoretical understanding, we point out two drawbacks in
GraphHop and further propose two corresponding enhancements, which lead
to an even more powerful semi-supervised solution called LERP. A
mathematical proof of its convergence is given. 
\item We demonstrate the effectiveness and efficiency of LERP with
extensive experiments on five commonly used datasets as well as an
object recognition task at extremely low label rates (i.e., 1, 2, 4, 8,
16 and 20 labeled samples per class). 
\end{enumerate}

The rest of this paper is organized as follows.  The problem definition
and some background information are stated in Sec.
\ref{sec:preliminaries}. A label efficient regularization and
propagation framework is set up and its connection with GraphHop is
built in Sec.  \ref{sec:graphhop}.  The LERP method is presented in Sec.
\ref{sec:lerp}. Theoretical analysis of the convergence and
computational complexity are analyzed in Sec.
\ref{sec:theoretical_analysis}. Extensive experiments are conducted for
performance benchmarking in Sec.  \ref{sec:experiments}. Comments on
related work are provided in Sec.  \ref{sec:related_work}. Finally,
concluding remarks are given in Sec.  \ref{sec:conclusion}. 

\section{Preliminaries}\label{sec:preliminaries}

\subsection{Notations and Problem Statement}

To begin with, we define some notations used throughout this paper and
give the definition to the investigated problem. An undirected graph is
given by $\mathcal{G}=(\mathcal{V}, \mathcal{E})$, where $\mathcal{V}$
is the node set with $|\mathcal{V}|=n$ and $\mathcal{E}$ is the edge
set. The graph structure can be described by adjacency matrix
$\mathbf{A}\in\mathbb{R}^{n\times n}$, where $\mathbf{A}_{ij}=1$ if node
$i$ and $j$ are linked by an edge in $\mathcal{G}$. Otherwise,
$\mathbf{A}_{ij}=0$. For $\mathbf{A}$, its diagonal degree matrix is
$\mathbf{D}_{ii}=\sum_{j=1}^{n}\mathbf{A}_{ij}$ and its graph Laplacian
matrix is $\mathbf{L}=\mathbf{D}-\mathbf{A}$. In an attributed graph,
each node is associated with a $d$-dimensional feature vector
$\mathbf{x}_i\in\mathbb{R}^d$, where the feature space of all nodes is
$\mathbf{X}\in\mathbb{R}^{n\times d}$. 

In a transductive semi-supervised classification problem, nodes in set
$\mathcal{V}$ can be divided into sets of labeled nodes $\mathcal{L}$
with $l$ samples and unlabeled nodes $\mathcal{U}$ with $u$ samples
followed by $n=l+u$. Let $\mathbf{Y}=(\mathbf{y}_1^T, \cdots,
\mathbf{y}_l^T, \mathbf{y}_{l+1}^T, \cdots, \mathbf{y}_n^T)^T
\in\mathbb{R}^{n\times c}$ denote the given labels of all samples,
where $\mathbf{y}_i\in\mathbb{R}^c$ is a row vector and $c$ is the
number of classes. If labeled node $i$ belongs to class $j$,
$\mathbf{y}_{ij} = 1$; otherwise, $\mathbf{y}_{ij}=0$. To record the
predicted label distribution of each node, the label embedding matrix is
defined as $\mathbf{F}=(\mathbf{f}_1^T, \cdots, \mathbf{f}_l^T,
\mathbf{f}_{l+1}^T, \cdots, \mathbf{f}_n^T)^T\in\mathbb{R}^{n\times c}$
with probability entries, where the $i$th row vector satisfying
$\sum_{j=1}^c\mathbf{f}_{ij}=1$ with $\mathbf{f}_{ij}$ is the
probability of node $i$ belonging to class $j$. The classes of unlabeled
nodes can be assigned as $y_i=\arg\max_{1\leq j \leq c}\mathbf{f}_{ij}$. 
Given graph $\mathcal{G}$, node attributes $\mathbf{X}$, and labeled
samples $\mathbf{Y}$, the objective is to infer labels of unlabeled
nodes under the condition $l\ll u$. 

\subsection{Transductive Label Propagation}

We introduce the general regularization framework for transductive LP
algorithms below. Formally, it can be defined as a problem that minimizes 
the objective function:
\begin{equation}\label{equ:regularization}
\min_\mathbf{F}\quad \text{tr}(\mathbf{F}^T\mathbf{\tilde{L}}\mathbf{F})
+ \text{tr}((\mathbf{F} - \mathbf{Y})^T\mathbf{U}(\mathbf{F} -\mathbf{Y})),
\end{equation}
where $\text{tr}(\cdot)$ is the trace operator, $\mathbf{\tilde{L}}$ is
the random-walk (or symmetric) normalized graph Laplacian matrix, and
$\mathbf{U}=\text{diag}(u_1, \cdots , u_n)$ is a positive weighting
matrix that balances the manifold smoothness and label fitness. For
generality, we relax the probability constraints on label embeddings
$\mathbf{F}$. 

The closed-form solution of Eq. (\ref{equ:regularization}) can be derived as
\begin{equation}\label{equ:lp_sol}
\mathbf{F}^* = (\mathbf{U} + \mathbf{\tilde{L}})^{-1}\mathbf{U}\mathbf{Y}.
\end{equation}
This result can also be obtained from an iteration process named LP.
At each iteration, the label embedding information of each node is partially
propagated by its neighbors and partially obtained from its given
label. Formally, at iteration $t$, the propagation can be described as
\begin{equation}\label{equ:lp}
\mathbf{F}^{(t)} = \alpha\mathbf{\tilde{A}}\mathbf{F}^{(t-1)} + (1 - \alpha)\mathbf{Y},
\end{equation}
where $\alpha\in(0, 1)$ is a weighting hyperparameter and
$\mathbf{\tilde{A}}$ is the random-walk (or symmetric) normalized
adjacency matrix. The converged label embedding matrix can be obtained by
taking the limit of iterations as
\begin{equation}\label{equ:lp_v2_sol}
\mathbf{F}^* = (1-\alpha)(\mathbf{I} - \alpha\mathbf{\tilde{A}})^{-1}\mathbf{Y},
\end{equation}
which is the same as Eq. (\ref{equ:lp_sol}) by setting
$$
\mathbf{U} = \text{diag}(\frac{1-\alpha}{\alpha}, \cdots, \frac{1-\alpha}{\alpha}).
$$

\subsection{GraphHop}

The recently proposed GraphHop model improves traditional LP methods and
offers state-of-the-art performance on several datasets. There are two
training stages in GraphHop (i.e., the initialization stage and the
iteration stage). They are used to capture the smooth node attributes
and label embeddings, respectively, as explained below.

In the initialization stage, smoothened attribute signals are
extracted by aggregating the neighborhood of each node. Formally, this
can be expressed as
\begin{equation}\label{equ:attribute_agg}
\mathbf{X}_M = ||_{0\leq m\leq M}\mathbf{\tilde{A}}^m\mathbf{X},
\end{equation}
where $||$ denotes column-wise concatenation, $\mathbf{\tilde{A}}^m$ is
the random-walk normalized $m$-hop adjacency matrix ($m=0$ indicates the
attribute of each node itself), and $\mathbf{X}_M\in\mathbb{R}^{n\times
d(M+1)}$ is the smoothened attribute matrix. After that, a logistic
regression (LR) classifier is adopted for training with labeled samples
as supervision. The minimization of cross-entropy loss can be written as
\begin{equation}\label{equ:graphhop_initialization}
\min_\mathbf{W}\quad -\frac{1}{l}\sum_{i=1}^l\mathbf{y}_i\log\left(
\sigma(\mathbf{W}\mathbf{x}_{M,i}^T)\right),
\end{equation}
where $\mathbf{x}_{M, i}$ is the $i$th row of $\mathbf{X}_M$,
$\mathbf{W}\in\mathbb{R}^{c \times d(M+1)}$ is the parameter matrix and
$\sigma(\mathbf{z})_i = e^{z_i}\big/\sum_{j=1}^ce^{z_j}$ is the softmax
function. The solution to Eq.  (\ref{equ:graphhop_initialization}) can
be derived by any optimization algorithm (e.g., stochastic gradient
descent). Once the classifier converges, the label embeddings of all
nodes can be predicted. They serve as the initial embeddings to the
subsequent iteration stage.  Formally, this can be written as
\begin{equation}\label{equ:attribute_infer}
\mathbf{F}^{(0)} = \mathbf{F}_{init} = \sigma(\mathbf{X}_M\mathbf{W}^T),
\end{equation}
where the softmax function, $\sigma(\mathbf{z})$, is applied in a
row-wise fashion. Note that in the original GraphHop model, distinct
classifiers are trained independently w.r.t. different hops of
aggregations, and the final results are the average of all predictions.
For simplicity, we only consider one classifier in the variational
derivation. The same applies to the iteration stage. 

In the iteration stage, there are two steps, called label aggregation
and label update, used for label embeddings propagation. 
In the label aggregation step, the same aggregation as given in Eq.
(\ref{equ:attribute_agg}) is conducted on label embeddings.
Formally, at iteration $t$, we have
\begin{equation}\label{equ:label_aggregation}
\mathbf{F}^{(t-1)}_M=||_{0\leq m\leq M}\mathbf{\tilde{A}}^m\mathbf{F}^{(t-1)},
\end{equation}
where $\mathbf{F}_M^{(t-1)}\in\mathbb{R}^{n\times c(M+1)}$ is the
aggregated nodes' neighborhood embeddings. In the aggregation of Eq.
(\ref{equ:label_aggregation}), the embedding parameters are not shared
between nodes, i.e., each label embedding is independently derived from
their neighborhoods. This will lead to deficient information passing
and deteriorate the performance, especially at low label rates. 
In the label update step, GraphHop trains an LR classifier on the
aggregated embedding space and uses inferred label embeddings at the
next iteration to overcome the shortcoming mentioned above. 
The minimization problem of the LR classifier can be stated as
\begin{equation}\label{equ:label_update_loss}
\begin{aligned}
\min_{\mathbf{W}}\quad  - & \sum_{i=1}^l\mathbf{y}_i\log\left(
\sigma(\mathbf{W}\mathbf{f}_{M, i}^{(t-1)^T})\right) \\
& - \alpha \sum_{i=l+1}^{n}\text{Sharpen}(\mathbf{f}_i^{(t-1)})
\log\left(\sigma(\mathbf{W}\mathbf{f}_{M, i}^{(t-1)^T})\right),
\end{aligned}
\end{equation}
where $\mathbf{f}_{M, i}^{(t-1)}$ (resp. $\mathbf{f}_i^{(t-1)}$) is the
$i$th row of $\mathbf{F}_M^{(t-1)}$ (resp. $\mathbf{F}^{(t-1)}$),
$\mathbf{W}\in\mathbb{R}^{c\times c(M+1)}$ is the parameter matrix,
$\alpha$ is a weighting hyperparameter, and
$$
\text{Sharpen}(\mathbf{z})_i=z_i^{\frac{1}{T}}\big/\sum_{j=1}^cz_j^{\frac{1}{T}}
$$
is the sharpening function to control the confidence of the current
label embedding distributions. The first term in the objective function
is contributed by labeled samples, while the second term is from
unlabeled samples, which are pseudo-labels generated by the classifier.
Note that we can view the classifier training as a smoothening operation
on label signals by sharing the same classifier parameters. Once 
Eq. (\ref{equ:label_update_loss}) has been solved, the inferred label 
embeddings at iteration $t$ are
\begin{equation}\label{equ:label_update}
\mathbf{F}^{(t)} = \sigma(\mathbf{F}_M^{(t-1)}\mathbf{W}^T).
\end{equation}
They serve as the input to the next iteration through Eq.
(\ref{equ:label_aggregation}). In the entire process of GraphHop, each
label embedding is a probability vector. 

In summary, the smoothened attributes on graphs are extracted and used
for label distribution prediction in the initialization stage. These
label embeddings are further smoothened by aggregation and classifier
training in the iteration stage and used for final inference. 

\section{Connection between GraphHop and LERP Framework}\label{sec:graphhop}

The superior performance of GraphHop was explained by joint smoothening
of node attributes and label embeddings through propagation and
classifier training in the last section. Here, we interpret the
smoothening process in GraphHop using a regularization framework,
show that it corresponds to an alternate optimization process of a
certain objective function with probability constraints. 

\subsection{Proposed Label Efficient Regularization and Propagation (LERP)
 Framework}

The initialization and iteration stages of GraphHop actually share a
similar procedure.  The main iterative process arises in the iteration
stage. In this subsection, we will derive a variational interpretation
for such an iterative process. To begin with, we set up a constrained
optimization problem, i.e., the LERP framework:
\begin{equation}\label{equ:graphhop_variational}
\begin{aligned} &
\begin{aligned}
\min_{\mathbf{F}, \mathbf{W}}\quad & \text{tr}(\mathbf{F}^T\mathbf{\tilde{L}}\mathbf{F}) 
+ \text{tr}\left((\mathbf{F}-\mathbf{F}_{init})^T\mathbf{U}(\mathbf{F}-\mathbf{F}_{init})\right)\\
& + \text{tr}\left((\mathbf{F}-\sigma(\mathbf{F}_M\mathbf{W}^T))^T\mathbf{U}_\alpha(\mathbf{F}
-\sigma(\mathbf{F}_M\mathbf{W}^T))\right)
\end{aligned} \\
& \text{s.t.}\quad \mathbf{F}_M=||_{0\leq m\leq M}\mathbf{\tilde{A}}^m\mathbf{F},\quad \mathbf{F} 
\mathbf{1}_c = \mathbf{1}_n,\quad \mathbf{F}\geq \mathbf{0}
\end{aligned}
\end{equation}
where $\mathbf{\tilde{L}}=\mathbf{D}^{-1}\mathbf{L}$ is the random-walk
normalized graph Laplacian matrix, $\mathbf{F}_{init}$ is the
initialized label embeddings from Eq. (\ref{equ:attribute_infer}),
$\sigma(\mathbf{z})$ is the softmax function applied row-wisely,
$\mathbf{U}\in\mathbb{R}^{n\times n}$ and
$\mathbf{U}_\alpha\in\mathbb{R}^{n\times n}$ are diagonal
hyperparameter matrix with positive entries,
$\mathbf{1}_c\in\mathbb{R}^c$ and $\mathbf{1}_n\in\mathbb{R}^{n}$ are
column vectors of ones, and $\mathbf{W}\in\mathbb{R}^{c\times c(M+1)}$
is the parameter matrix. The last two constraints make each label
embedding a probability vector. The cost function in Eq.
(\ref{equ:graphhop_variational}) consists of three terms.  The first two
give the objective function of the general regularization framework in
Eq. (\ref{equ:regularization}).  The third term is the Frobenius norm on
$\mathbf{F}-\sigma(\mathbf{F}_M\mathbf{W}^T)$ if $\mathbf{U}_\alpha$ is
the identity matrix. It can be viewed as further regularization.  The
functionality of this extra regularization term can be seen as forcing
the label embeddings to be close to the classifier predictions based on
neighborhood representations. Thus, the embedding information around
labeled nodes can be effectively passed farther distances to unlabeled
nodes by sharing a similar neighborhood. This enables the LERP method
to have superior performance in extremely low label rates as discussed later in
the optimization process. Note that we have replaced the matrix
$\mathbf{Y}$ with the initialized label embeddings, $\mathbf{F}_{init}$,
derived from Eq. (\ref{equ:attribute_infer}) in the cost function. 

\subsection{Alternate Optimization}

We present the optimization procedure for the solution of Eq.
(\ref{equ:graphhop_variational}) and discuss its relationship with
GraphHop. Due to the introduction of variable $\mathbf{W}$ in the extra
regularization term, the optimization of two variables, $\mathbf{F}$ and
$\mathbf{W}$, depends on each other. The problem cannot be solved
directly but alternately. We propose an alternate optimization strategy
that updates one variable at a time. This alternate optimization
strategy has two subproblems. Each of them is a non-convex optimization
problem. With approximation and convex transformation, we show that they
correspond to the label aggregation and the label update steps in the
iteration stage of GraphHop, respectively. 

\subsubsection{Updating $\mathbf{W}$ with Fixed $\mathbf{F}$}
\label{sec:optimization_classifier}

The classifier parameters, $\mathbf{W}$, only exist in the extra
regularization term of Eq. (\ref{equ:graphhop_variational}).
Thus, the optimization problem is
\begin{equation}\label{equ:graphhop_optimize_w}
\min_{\mathbf{W}}\quad \text{tr}\left((\mathbf{F}-\sigma(\mathbf{F}_M 
\mathbf{W}^T))^T\mathbf{U}_\alpha(\mathbf{F}-\sigma(\mathbf{F}_M\mathbf{W}^T))\right).
\end{equation}
By setting $\mathbf{U}_\alpha=\text{diag}(u_{\alpha, 1}, \cdots,
u_{\alpha, n})$, Eq. (\ref{equ:graphhop_optimize_w}) can be written as
\begin{equation}\label{equ:graphhop_optimize_w_2}
\min_{\mathbf{W}}\quad \sum_{i=1}^nu_{\alpha, i}||\mathbf{f}_i^T - 
\sigma(\mathbf{W}\mathbf{f}_{M, i}^T)||_2^2,
\end{equation}
which is a weighted sum of the squared $l_2$-norm on the difference between each
label embedding and the classifier prediction. It is a non-convex problem
due to the softmax function.  To address it, we consider the following
transformation. Since $\mathbf{f}_i$ and $\sigma(\mathbf{W}
\mathbf{f}_{M, i}^T)$ are both probability vectors, it is more
appropriate to use the KL-divergence to measure their distance.  Then,
by converting the $l_2$-norm to the KL-divergence and only considering
the term w.r.t. parameter $\mathbf{W}$, we are led to the cross-entropy
loss minimization:
\begin{equation}\label{equ:graphhop_optimize_w_cross_entropy}
\min_{\mathbf{W}}\quad -\sum_{i=1}^{n}u_{\alpha, i}\mathbf{f}_i 
\log(\sigma(\mathbf{W}\mathbf{f}_{M, i}^T)).
\end{equation}
Note that the optimal solutions to Eq. (\ref{equ:graphhop_optimize_w_2})
and Eq.  (\ref{equ:graphhop_optimize_w_cross_entropy}) are identical.
Besides, we have the following Theorem. 
\begin{theorem}\label{theorem:convex_problem_w}
    The optimization problem in Eq.
    (\ref{equ:graphhop_optimize_w_cross_entropy}) is a convex optimization
    problem. 
\end{theorem}
The proof is given in Appendix \ref{app:convex_problem_w}. 

To further reveal the connection between GraphHop, we rewrite the
cost function as contributions from labeled and unlabeled samples in the
form of
\begin{equation}\label{equ:graphhop_optimize_w_cross_entropy_2}
-\sum_{i=1}^{l}u_{\alpha, i}\mathbf{f}_i\log(\sigma(\mathbf{W}\mathbf{f}_{M, i}^T))
-\sum_{i=l+1}^{n}u_{\alpha, i}\mathbf{f}_i\log(\sigma(\mathbf{W}\mathbf{f}_{M, i}^T)).
\end{equation}
Instead of adopting the current label embedding (i.e., $\mathbf{f}_i$)
as supervision for distribution mapping directly, two improvements can
be made to the labeled and unlabeled terms, respectively. For the
labeled term, we can replace label embeddings with ground-truth
labels (i.e., $\mathbf{y}_i$) for supervision. For the unlabeled term,
since there is no ground-truth label and the current supervision is
adopted from the former iteration (i.e., pseudo-labels), a sharpening
function can be used to control the confidence of the present
pseudo-labels for supervision. Also, by simplifying weighting
hyperparameter $u_{\alpha, i}$ to the same value (i.e., $1$ for labeled data
and $\alpha$ for unlabeled data), Eq.
(\ref{equ:graphhop_optimize_w_cross_entropy_2}) becomes
\begin{equation}\label{equ:label_update_loss_2}
\begin{aligned}
\min_{\mathbf{W}}\quad - & \sum_{i=1}^l  \mathbf{y}_i\log\left(
\sigma(\mathbf{W}\mathbf{f}_{M, i}^{T})\right) \\
 & - \alpha \sum_{i=l+1}^{n}\text{Sharpen}(\mathbf{f}_i)\log
\left(\sigma(\mathbf{W}\mathbf{f}_{M, i}^{T})\right),
\end{aligned}
\end{equation}
which is exactly the objective function in the label update step of
GraphHop as given by Eq.  (\ref{equ:label_update_loss}). Eq.
(\ref{equ:label_update_loss_2}) is a convex function, and its optimum
can be obtained by a standard optimization procedure. 

\subsubsection{Updating $\mathbf{F}$ with Fixed $\mathbf{W}$} 
\label{sec:optimization_embedding}

With fixed classifier parameters $\mathbf{W}$, the optimization problem
in Eq. (\ref{equ:graphhop_variational}) becomes
\begin{equation}\label{equ:graphhop_optimize_f}
\begin{aligned} &
\begin{aligned}
\min_{\mathbf{F}}\quad & \text{tr}(\mathbf{F}^T\mathbf{\tilde{L}}\mathbf{F}) 
+ \text{tr}((\mathbf{F}-\mathbf{F}_{init})^T\mathbf{U}(\mathbf{F}-\mathbf{F}_{init}))\\
& + \text{tr}\left((\mathbf{F}-\sigma(\mathbf{F}_M\mathbf{W}^T))^T\mathbf{U}_\alpha(
\mathbf{F}-\sigma(\mathbf{F}_M\mathbf{W}^T))\right)
\end{aligned} \\
& \text{s.t.}\quad \mathbf{F}_M=||_{0\leq m\leq M}\mathbf{\tilde{A}}^m\mathbf{F},\quad 
\mathbf{F}\mathbf{1}_c = \mathbf{1}_n,\quad \mathbf{F}\geq \mathbf{0}.
\end{aligned}
\end{equation}
This is a non-convex optimization problem due to the last regularization
term which involves both parameters $\mathbf{F}$ and $\mathbf{F}_M$.  To
solve this, we argue that $\mathbf{F}$ and $\mathbf{F}_M$ in the last
term should not be optimized simultaneously. The reason is that, in
deriving the former subproblem in Eq.  (\ref{equ:label_update_loss_2}),
the supervision of unlabeled samples are the pseudo-labels from previous
iterations. The pseudo-labels should serve as the input (rather than
parameters) in the current optimization. For simplicity, we fix the
parameter inside $\mathbf{F}_M$\footnote{$\mathbf{F}$ and
$\sigma(\mathbf{F}_M\mathbf{W})$ are initially the same due to the
optimization in Eq.  (\ref{equ:label_update_loss_2})).}.
Then, we obtain the following optimization problem:
\begin{equation}\label{equ:graphhop_optimize_f_fixed}
\begin{aligned} &
\begin{aligned}
\min_{\mathbf{F}}\quad & \text{tr}(\mathbf{F}^T\mathbf{\tilde{L}}\mathbf{F}) 
+ \text{tr}((\mathbf{F}-\mathbf{F}_{init})^T\mathbf{U}(\mathbf{F}-\mathbf{F}_{init}))\\
& + \text{tr}\left((\mathbf{F}-\sigma(\mathbf{F}_M\mathbf{W}^T))^T\mathbf{U}_\alpha(
\mathbf{F}-\sigma(\mathbf{F}_M\mathbf{W}^T))\right)
\end{aligned} \\
& \text{s.t.}\quad \mathbf{F}_M=||_{0\leq m\leq M}\mathbf{\tilde{A}}^m\mathbf{\hat{F}},
\quad \mathbf{F}\mathbf{1}_c = \mathbf{1}_n,\quad \mathbf{F}\geq \mathbf{0},
\end{aligned}
\end{equation}
where $\mathbf{F}_M$ is now a constant derived from the starting label embeddings $\mathbf{\hat{F}}$ of this optimization subproblem. We have the following
theorem for this problem. 
\begin{theorem}\label{theorem:convex_problem}
The problem in Eq. (\ref{equ:graphhop_optimize_f_fixed}) is 
a convex optimization problem.
\end{theorem}
The proof can be found in Appendix \ref{app:convex_problem}. 

A straightforward way to solve this problem is to apply the KKT
conditions \cite{boyd2004convex} to the Lagrangian function. Instead, we
show a different derivation which is intuitively connected to
GraphHop and leading to the same optimum. First, we directly take the derivative 
of the cost function and set it to zero. It gives
\begin{equation}\label{equ:optimization_solution}
\mathbf{F}^* = (\mathbf{U} + \mathbf{U}_\alpha + \mathbf{\tilde{L}})^{-1} 
(\mathbf{U}\mathbf{F}_{init}+\mathbf{U}_\alpha\sigma(\mathbf{F}_M\mathbf{W}^T)).
\end{equation}
Next, we introduce two constants $\mathbf{U}^\prime$ and $\mathbf{Y}^\prime$:
\begin{equation}\label{equ:variables}
\begin{cases}
\begin{aligned}
\mathbf{U}^\prime &= \mathbf{U} + \mathbf{U}_\alpha \\
\mathbf{U}^\prime\mathbf{Y}^\prime &= \mathbf{U}\mathbf{F}_{init} 
+ \mathbf{U}_\alpha\sigma(\mathbf{F}_M\mathbf{W}^T))
\end{aligned}
\end{cases}.
\end{equation}
By substituting these two terms in Eq. (\ref{equ:optimization_solution}), 
we obtain
\begin{equation}\label{equ:optimization_solution_2}
\mathbf{F}^* = (\mathbf{U}^\prime + \mathbf{\tilde{L}})^{-1} 
\mathbf{U}^\prime\mathbf{Y}^\prime.
\end{equation}
It has the same form as Eq. (\ref{equ:lp_sol}). Thus, the same result
can be derived from the label propagation process as given in Eq.
(\ref{equ:lp}). It can be expressed as
\begin{equation}\label{equ:variational_lp}
\resizebox{.99\hsize}{!}{$
\begin{aligned} 
\mathbf{F}^{(t)} & = \mathbf{U}_\beta\mathbf{\tilde{A}}\mathbf{F}^{(t-1)} 
+ (\mathbf{I} - \mathbf{U}_\beta)\mathbf{Y}^\prime \\
 &\begin{aligned} 
  \ = &\  \mathbf{U}_\beta\mathbf{\tilde{A}}\mathbf{F}^{(t-1)}+(\mathbf{I} - \mathbf{U}_\beta) \\
  &\times\big((\mathbf{U} + \mathbf{U}_\alpha)^{-1}\mathbf{U}\mathbf{F}_{init}
  + (\mathbf{U}+\mathbf{U}_\alpha)^{-1}\mathbf{U}_\alpha\sigma(\mathbf{F}_M\mathbf{W}^T)\big)
\end{aligned}
\end{aligned}$},
\end{equation}
where 
$$\mathbf{U}_\beta=(\mathbf{I} + \mathbf{U}^\prime)^{-1} =
(\mathbf{I} + \mathbf{U} + \mathbf{U}_\alpha)^{-1}. 
$$
This is summarized in the following Proposition. 
\begin{proposition}\label{proposition}
The iteration process in Eq. (\ref{equ:variational_lp}) converges to Eq.
(\ref{equ:optimization_solution_2}) and further to Eq.
(\ref{equ:optimization_solution}). 
\end{proposition}
The proof is given in Appendix \ref{app:proposition}. Also, we have the
following theorem stating the relationships between
the convergence result, i.e., Eq.  (\ref{equ:optimization_solution}),
and the optimal solution to the optimization problem in Eq.
(\ref{equ:graphhop_optimize_f_fixed}). 
\begin{theorem}\label{theorem:iteration_optimal}
If the variable, $\mathbf{F}$, is started in a probabilistic way,
i.e., $\mathbf{F}^{(0)}\mathbf{1}_c=\mathbf{1}_n$ and
$\mathbf{F}^{(0)}\geq\mathbf{0}$, then the convergence result (i.e., 
Eq. (\ref{equ:optimization_solution})) of the
iteration process in Eq. (\ref{equ:variational_lp}) is the optimal
solution to the optimization problem in Eq.
(\ref{equ:graphhop_optimize_f_fixed}). 
\end{theorem}
The proof is shown in Appendix \ref{app:iteration_optimal}. 

Note that starting with a probabilistic distribution can be easily achieved. That
is, we can directly adopt the result from the former round of
optimization and initialize it as $\mathbf{F}_{init}$. Theorem
\ref{theorem:iteration_optimal} provides an iterative solution to the
optimization problem (\ref{equ:graphhop_optimize_f_fixed}). By comparing
the iteration process in Eq. (\ref{equ:variational_lp}) with the label
aggregation in Eq.  (\ref{equ:label_aggregation}) and the label update
in Eq. (\ref{equ:label_update}) of GraphHop, we see that GraphHop only
uses the last term of Eq. (\ref{equ:variational_lp}) without iterations,
i.e.,
\begin{equation}\label{equ:graphhop_embedding_update}
\mathbf{F}^*\approx \mathbf{F}^{(0)} = \sigma(\mathbf{F}_M\mathbf{W}^T), \quad \text{where}\ 
\mathbf{F}_M=||_{0\leq m\leq M}\mathbf{\tilde{A}}^m\mathbf{\hat{F}}.
\end{equation}
This is a rough approximation to the iteration process in Eq.
(\ref{equ:variational_lp}), as well as the optimum in Eq.
(\ref{equ:optimization_solution}). On the one hand, it does not yield the
optimal solution to Eq.  (\ref{equ:graphhop_optimize_f_fixed}) in
general. On the other hand, it still meets the probability constraints.
Most importantly, it lowers the cost of optimizing Eq.
(\ref{equ:graphhop_optimize_f_fixed}), which makes GraphHop scalable to
large-scale networks. 

We summarize the main results of this section as follows. GraphHop
offers an alternate optimization solution to the variational problem
(\ref{equ:graphhop_variational}). It roughly solves two approximate subproblems
alternatively and iteratively. Its classifier training in the label
update step is a convex transformation of the optimization problem in
Eq.  (\ref{equ:graphhop_optimize_w}). Its label aggregation and update
steps as given in Eqs.  (\ref{equ:label_aggregation}) and
(\ref{equ:label_update}) yield an approximate solution to the
optimization problem in Eq.  (\ref{equ:graphhop_optimize_f_fixed}),
which itself is an approximated convex transformation of the problem in
Eq.  (\ref{equ:graphhop_optimize_f}).  Based on this understanding, we
propose an enhancement, named LERP, to address the limitations of
GraphHop. 

\section{Proposed LERP Method}\label{sec:lerp}

\begin{figure}[!t]
\centering
\includegraphics[width=0.48\textwidth]{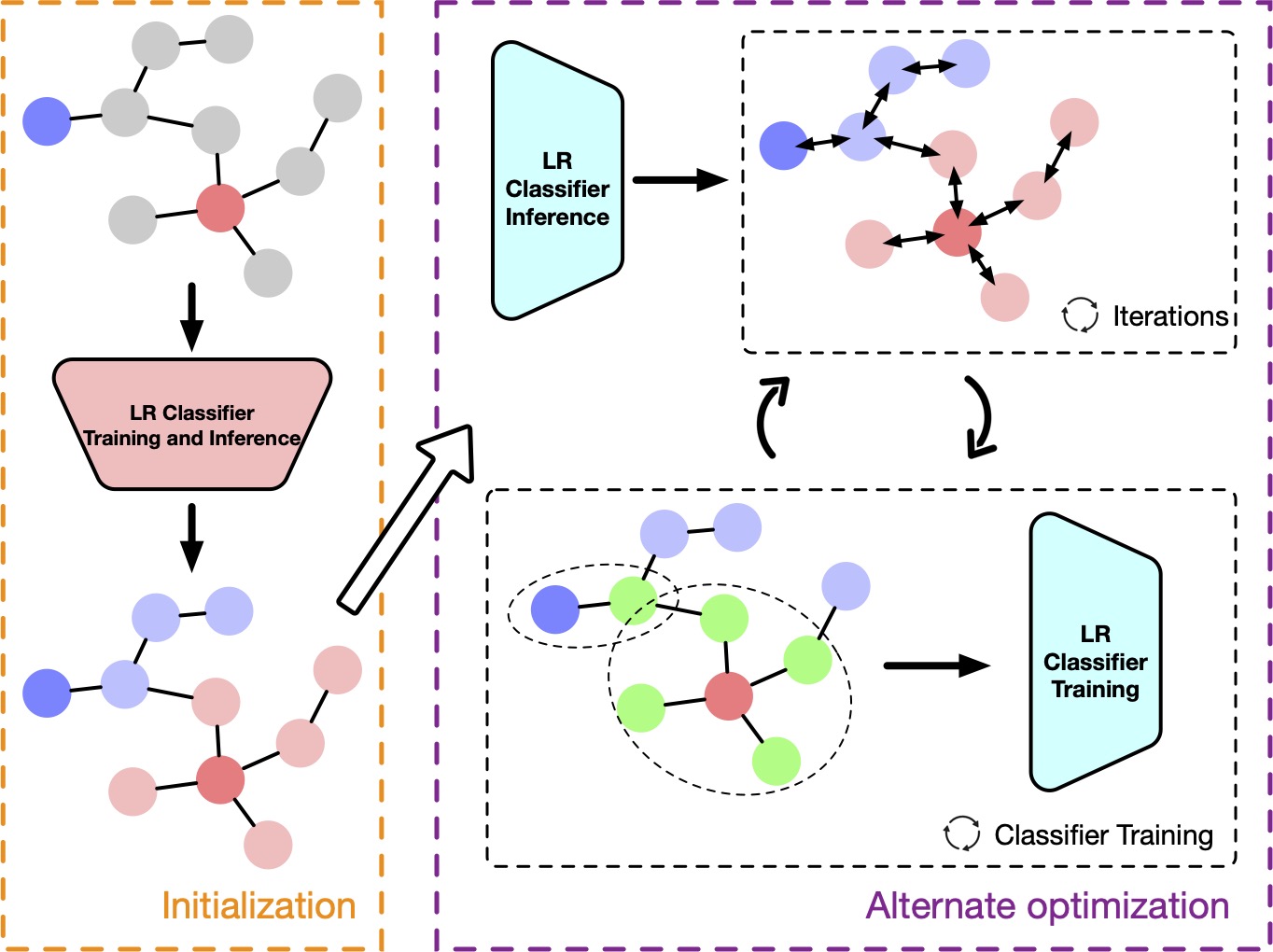}
\caption{An overview of the LERP method, where the left subfigure
shows the initialization for the subsequent process while the right
subfigure depicts the alternate optimization process.  The latter
consists of: 1) iterations for label embeddings update and 2) classifier
training for classifier parameters update.}\label{fig:graphhop_plus}
\end{figure}

The discussion in the last section reveals two drawbacks in GraphHop,
which also leads to two ideas to their improvements. They are elaborated in
Secs.  \ref{subsec:enhancement_1} and \ref{subsec:enhancement_2},
respectively.  An overview of the LERP model is shown in Fig.
\ref{fig:graphhop_plus}. In the left subfigure, its upper part shows
labeled nodes from two different classes (in blue and red) and unlabeled
nodes (in gray), while its lower part shows label embeddings predicted
by the LR classifier. They serve as the initialization for the
subsequent alternate optimization process.  The right subfigure depicts
the alternate optimization process. It consists of two alternate
optimization steps: 1) iterations for label embeddings update and 2)
classifier training for classifier parameters update.  In the right
subfigure, nodes in the curriculum set are colored in green.  Only the
labeled nodes and nodes in the curriculum set are employed in the
classifier training step. 

%%%%%%%%%%%%%%%%%%%%%%%%%%%%%%%%%%%%%%%%%%%%%%%%%%%%%%%%%%%%%%%
%%%%%%%%%%%%%%%%%%%%%%%%%%%%%%%%%%%%%%%%%%%%%%%%%%%%%%%%%%%%%%%

%%%%%%%%%%%%%%%%%%%%%%%%%%%%%%%%%%%%%%%%%%%%%%%%%%%%%%%%%%%%%%%
\begin{figure}[!t]
\centering
\includegraphics[width=0.48\textwidth]{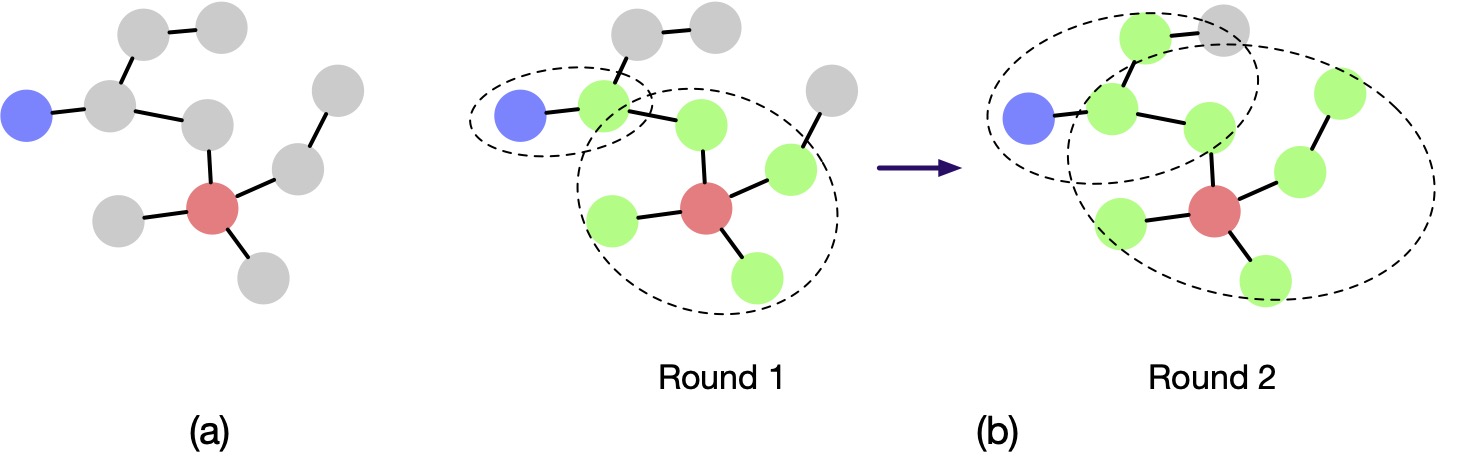}
\caption{Illustration of growing curriculum sets in the alternate
optimization process: (a) the original graph, (b) and (c) selected
reliable nodes in the curriculum set in the first and second rounds of
the optimization process. Two labeled nodes of two classes are colored
in blue and red, unlabeled nodes and nodes in the curriculum set are
colored in gray and green, respectively.  The dotted ellipses show the
corresponding curriculum sets of two labeled nodes.}
\label{fig:enhancement_classifier_training}
\end{figure}
%%%%%%%%%%%%%%%%%%%%%%%%%%%%%%%%%%%%%%%%%%%%%%%%%%%%%%%%%%%%%%%

\subsection{Enhancement in Classifier Training}\label{subsec:enhancement_1}

In the subproblem of classifier parameters updating with fixed label
embeddings as described in Sec. \ref{sec:optimization_classifier}, the
substituted objective function in GraphHop is given in Eq.
(\ref{equ:label_update_loss_2}).  Unlabeled samples are supervised by
label embeddings from the previous iteration, which can be regarded as a
self-training process with pseudo-labels. However, the pseudo-labels may
not be reliable, which may have a negative impact on classifier
training. Besides, this self-training may suffer from label error
feedback \cite{lee2013pseudo}, which makes the classifier biased in the
wrong direction. Thus, we should not include all unlabeled samples but
the most trustworthy ones in classifier training. Intuitively, the
reliability of pseudo-labels of unlabeled samples can be measured by
their distances to labeled ones in the network; namely, the closer to
labeled samples, the more reliable the pseudo-labels. The rationale is
that ground-truth labels should have a stronger influence on their
neighborhoods than nodes farther away through propagation. 

Formally, we define the collection of reliable nodes as follows.
\begin{definition}\label{definition} 
The set of reliable nodes is defined by the curriculum set $\mathcal{S}$, 
which is a subset of unlabeled samples $\mathcal{S}\subseteq\mathcal{U}$. 
At the $r$th round (starting from $r=1$) of the alternate
optimization process, nodes in this set satisfy
$\mathcal{S}_{r}=\{u|\min_{v\in\mathcal{V}}\text{dist}(u, v)\leq r,
u\in\mathcal{U}\}$, where $\text{dist}(u, v)$ is the shortest
path between the two nodes.  
\end{definition}
In words, at the $r$th round of the optimization process, unlabeled
nodes that are within $r$-hop away from the labeled nodes are accepted
as reliable nodes and included in the curriculum set. This is illustrated
in Fig. \ref{fig:enhancement_classifier_training}. 

Then, we can modify the objective function of classifier training
in Eq. (\ref{equ:label_update_loss_2}) as
\begin{equation}\label{equ:graphhop_plus_classifier_loss}
\begin{aligned}
\min_{\mathbf{W}}\quad - & \sum_{i=1}^l  \mathbf{y}_i\log\left(
\sigma(\mathbf{W}\mathbf{f}_{M, i}^{T})\right) \\
& - \alpha \sum_{i\in\mathcal{S}_r}\text{Sharpen}(\mathbf{f}_i)\log
\left(\sigma(\mathbf{W}\mathbf{f}_{M, i}^{T})\right).
\end{aligned}
\end{equation}
As a result, only reliable nodes in the curriculum set contribute to
classifier training.  This enhancement also reduces the training time
since there are fewer noisy samples. 

\subsection{Enhancement in Label Embeddings Update}\label{subsec:enhancement_2}

For the subproblem of label embeddings update with fixed classifier
parameters as presented in Sec. \ref{sec:optimization_embedding}, we
showed that, starting with a probabilistic distribution of the label
embeddings, the optimal solution to Eq.
(\ref{equ:graphhop_optimize_f_fixed}) can be derived from an iteration
process as given in Eq. (\ref{equ:variational_lp}).  For GraphHop, it is
further simplified to only the predictions from the classifier without
iteration as given in Eq. (\ref{equ:graphhop_embedding_update}). It is,
however, a rough approximation.  GraphHop can be enhanced by considering
the optimal solution in Eq.  (\ref{equ:optimization_solution}). Still,
the inverse of the Laplacian matrix in Eq.
(\ref{equ:optimization_solution}) has high time complexity -
$\mathcal{O}(n^3)$, which is not practical. Instead, we adopt the
iteration in Eq.  (\ref{equ:variational_lp}) to approximate the optimal
solution with fewer iterations and avoid matrix inversion.
Specifically, we propose the following iteration:
\begin{equation}\label{equ:graphhop_plus_iteration}
\begin{aligned}
\mathbf{F}^{(t)} & = \mathbf{U}_\beta\mathbf{\tilde{A}}\mathbf{F}^{(t-1)} 
+ (\mathbf{I} - \mathbf{U}_\beta)\mathbf{F}_{init} \\
& = \beta\mathbf{\tilde{A}}\mathbf{F}^{(t-1)} + (1-\beta)\mathbf{F}_{init},
\end{aligned}
\end{equation}
where hyperparameters in $\mathbf{U}_\beta$ are simplified to be the
same $\beta\in(0, 1)$ for all nodes.  The starting value of the
iteration is given by
\begin{equation}\label{equ:graphhop_plus_iteration_init}
\mathbf{F}^{(0)} = \sigma(\mathbf{F}_M\mathbf{W}^T),\quad \text{where}\ 
\mathbf{F}_M=||_{0\leq m\leq M}\mathbf{\tilde{A}}^m\mathbf{\hat{F}}.
\end{equation}
Note that we drop the last classifier inference term
$\sigma(\mathbf{F}_M\mathbf{W}^T)$ in Eq.
(\ref{equ:graphhop_plus_iteration}) for simplicity since it has
already been included in Eq.  (\ref{equ:graphhop_plus_iteration_init}).
The balance between $\mathbf{F}_{init}$ and $\sigma(\mathbf{F}_M\mathbf{W}^T)$ can
also be achieved by adjusting the weight $\beta$.

If there is no iteration in Eq.  (\ref{equ:graphhop_plus_iteration})
(i.e., $t=0$), it degenerates to GraphHop. Another advantage of the
iteration is that it can alleviate the oversmoothening problem introduced
by continuous label embeddings update in Eq.
(\ref{equ:graphhop_embedding_update}). Since the initial label
embeddings $\mathbf{F}_{init}$ are added at each iteration as shown in Eq.
(\ref{equ:graphhop_plus_iteration}), the distinct embedding information
from the initialization stage is consistently enforced at each node
during the optimization process. 

With the above two enhancements, the pseudo-codes of LERP are given in
Algorithm \ref{alg:graphhop_plus}.  Note that we adopt the same label
initialization as GraphHop as given in Eq.
(\ref{equ:graphhop_initialization}), where the enhancements in the
alternate optimization are shown in lines \ref{alg:alternate_start} to
\ref{alg:alternate_end}. The LR classifier is a linear classifier by
nature. Yet, nonlinearity can be introduced with nonlinear kernels. This
has not yet been tried in this work. 

%%%%%%%%%%%%%%%%%%%%%%%%%%%%%%%%%%%%%%%%%%%%%%%%%%%%%%%%%%%%%%
\begin{algorithm}[!t]
\caption{LERP}\label{alg:graphhop_plus}
\begin{algorithmic}[1]
    \STATE \textbf{Input}: Graph $\mathbf{A}$, attributes $\mathbf{X}$, label vectors $\mathbf{Y}$
    \STATE \textbf{Initialization}:
    \STATE $\mathbf{X}_M\gets$ compute Eq. (\ref{equ:attribute_agg})
    \WHILE{not converged}
        \FOR{each minibatch of labeled nodes} \label{alg_line:batch_1}
            \STATE Compute $\mathbf{g}\gets\nabla{L}(\mathbf{X}_{M}, \mathbf{Y};\mathbf{W})$ in Eq. (\ref{equ:graphhop_initialization})
            \STATE Conduct Adam update using gradient estimator $\mathbf{g}$
        \ENDFOR
    \ENDWHILE
    \STATE $\mathbf{F}^{(0)} \gets$ compute Eq. (\ref{equ:attribute_infer}) 
    \STATE \textbf{Alternate optimization of Eq. (\ref{equ:graphhop_variational})}:
    \FOR{$r \in [1, ..., max\_round]$}
        \FOR{$t \in [1, ..., max\_iter]$}\label{alg:alternate_start}
            \STATE Compute Eq. (\ref{equ:graphhop_plus_iteration})
        \ENDFOR
        \STATE Update the curriculum set $\mathcal{S}$ defined in Definition \ref{definition}
        \WHILE{not converged}
            \FOR{each minibatch of all nodes} \label{alg_line:batch_2}
                \STATE Compute $\mathbf{g}\gets\nabla{L}(\mathbf{F}, \mathbf{F}_M, \mathbf{Y}; \mathbf{W})$ in Eq. (\ref{equ:graphhop_plus_classifier_loss})
                \STATE Conduct Adam update using gradient estimator $\mathbf{g}$
            \ENDFOR
        \ENDWHILE
        \STATE Compute Eq. (\ref{equ:graphhop_plus_iteration_init})\label{alg:alternate_end}
    \ENDFOR
    \STATE \textbf{Output}: Labels of unlabeled nodes $y_i = \arg\max_{1\leq j\leq c}\mathbf{f}_{ij}$
\end{algorithmic}
\end{algorithm}
%%%%%%%%%%%%%%%%%%%%%%%%%%%%%%%%%%%%%%%%%%%%%%%%%%%%%%%%%%%%%%

\section{Theoretical Analysis}\label{sec:theoretical_analysis}

In this section, we give the convergence analysis and complexity
analysis of the LERP method. 

\subsection{Convergence Analysis}

\begin{theorem}\label{theorem:convergence}
The LERP method monotonically decreases the value of the cost function
(\ref{equ:graphhop_variational}) within a constant difference
$2\emph{tr}(\mathbf{U}_\alpha)$ and with constraints satisfied in each
round until the algorithm converges. 
\end{theorem}
The proof is shown in Appendix \ref{app:convergence}.

Theorem \ref{theorem:convergence} indicates that LERP method guarantees
the convergence of cost function in problem
(\ref{equ:graphhop_variational}) with all constraints satisfied.
Experimental results show that LERP usually converges in less than 20
alternate rounds while the number of iterations of Eq.
(\ref{equ:graphhop_plus_iteration}) is less than 10. 

\subsection{Complexity Analysis}

The analysis of time and memory complexities of LERP is similar to
\cite{xie2022graphhop}. Specifically, we consider one minibatch of nodes
since every step in Algorithm \ref{alg:graphhop_plus} can be exercised
accordingly. Assume that, during optimization, the number of minibatches
is $N$, the number of nodes is $n$, the alternate round of optimization
is $r$, the number of iterations is $t$, the size of one minibatch is
$b$, and the number of classes is $c$. Then, the time complexity of one
minibatch during the entire optimization can be computed as
$\mathcal{O}\left(r(n||\mathbf{\tilde{A}}^M||_0 +
c||\mathbf{\tilde{A}}^M||_0/N + tc||\mathbf{\tilde{A}}||_0/N +
tbc^2)\right)$ where the first and the second term come from the
computation of $M$-hop neighbors and the corresponding aggregation of
label embeddings in Eq. (\ref{equ:graphhop_plus_iteration_init}), the
third term comes from the iteration in Eq.
(\ref{equ:graphhop_plus_iteration}), and the last term derives from the
classifier training in Eq. (\ref{equ:graphhop_plus_classifier_loss}).
Note that we can eliminate the first term by only considering the
one-hop neighbors (i.e., $M = 1$). 

The memory usage complexity can be expressed as $\mathcal{O}(bc + c^2)$
which represents one minibatch embedding and the parameters of the
classifier. We ignore the storage of the adjacency matrix since it is
the same for all algorithms. Note that the memory cost scales linearly
in terms of the minibatch size $b$ and is fixed during the entire
optimization process. 

\section{Experiments}\label{sec:experiments}

We evaluate LERP on the transductive semi-supervised node
classification task with multiple real-word graph datasets in Secs.
\ref{subsec:datasets}-\ref{subsec:ablation}.  Furthermore, we apply it
to an object recognition problem in Sec.
\ref{subsec:object_recognition}

\subsection{Datasets}\label{subsec:datasets}

For performance benchmarking, we consider five widely used graph
datasets whose statistics are shown in Table \ref{tab:datasets},
including the numbers of nodes, edges, node labels (i.e., classes), and
node feature dimensions. Among them, Cora, CiteSeer, and PubMed
\cite{sen2008collective, yang2016revisiting, kipf2016semi} are three
citation networks. Their nodes represent documents, links indicate
citations between documents, labels denote document's categories, and
node features are bag-of-words vectors. Amazon Photo
\cite{mcauley2015image, shchur2018pitfalls} is a co-purchase network.
Its nodes represent goods items, and links suggest that the two goods
items are frequently purchased together. Its labels are product
categories, and node features are bag-of-words encoded product reviews.
Coauthor CS \cite{shchur2018pitfalls} is a co-authorship graph.  Its
nodes are authors, which are connected by an edge if they have a
co-authored paper.  Node features are keywords of authors' papers, while
a node label indicates the author's most active fields of study. 

%%%%%%%%%%%%%%%%%%%%%%%%%%%%%%%%%%%%%%%%%%%%%%%%%%%%%%%%%%%%%%
\begin{table}[!t]
\renewcommand{\arraystretch}{.9}
\caption{Benchmark dataset properties and statistics.}
\label{tab:datasets}
\centering
\begin{tabular}{ccccc} \hline
    \textbf{Dataset} & \textbf{Nodes} & \textbf{Edges} & \textbf{Classes} 
    & \textbf{Feature Dims} \\ \hline
    Cora & $2,708$ & $5,429$ & $7$ & $1,433$ \\
    CiteSeer & $3,327$ & $4,732$ & $6$ & $3,703$ \\
    PubMed & $19,717$ & $44,338$ & $3$ & $500$ \\ 
    Amazon Photo & $7,487$ & $119,043$ & $8$ & $745$ \\ 
    Coauthor CS & $18,333$ & $81,894$ & $15$ & $6,805$ \\ \hline
\end{tabular}
\end{table}
%%%%%%%%%%%%%%%%%%%%%%%%%%%%%%%%%%%%%%%%%%%%%%%%%%%%%%%%%%%%%%

For data splitting, the standard evaluation of semi-supervised node
classification is to sample 20 labeled nodes per class as the training
set \cite{kipf2016semi, shchur2018pitfalls}. To demonstrate the label
efficiency of LERP, we examine extremely low label rates in the
training set. That is, we sample 1, 2, 4, 8, 16, and 20 labeled nodes
per class randomly to form the training set. After that, we follow the
standard evaluation procedure by using 500 of the remaining samples as
the validation set and the rest as the test dataset. We apply the same
dataset split rule to all benchmarking methods for a fair comparison. 

\subsection{Experimental Settings}\label{subsec:setting}

\subsubsection{Benchmarking Methods, Performance Metrics and
Experimental Environment} We compare LERP with representative
state-of-the-art methods of the following three categories. 
\begin{enumerate}
\item GCN-based Methods: GCN \cite{kipf2016semi} and GAT
\cite{velickovic2018graph}. 
\item Propagation-based Methods: LP \cite{zhou2003learning}, APPNP
\cite{klicpera2018predict}, C\&S \cite{huang2020combining}, and GraphHop
\cite{xie2022graphhop}. 
\item Label-efficient GCN-based Methods: Co-training GCN
\cite{li2018deeper}, self-training GCN \cite{li2018deeper}, IGCN
\cite{li2019label}, GLP \cite{li2019label}, and CGPN
\cite{wan2021contrastive}. 
\end{enumerate}
We implement GCN and GAT using the PyG library \cite{Fey/Lenssen/2019}
and LP based on its description. For other benchmarking methods, we
adopt their released code. All experimental settings are the same as
those specified in their original papers.  We run each method on the
five datasets as described in Sec.  \ref{subsec:datasets}. 

All experiments for LERP and benchmarking methods are conducted
ten times on each dataset. The mean test accuracy and the standard
deviation are used as the evaluation metric. The experimental
environment is a machine with an NVIDIA Tesla V100 GPU (32-GB memory), a
ten-core Intel Xeon CPU and 50 GB of RAM. 

\subsubsection{Implementation Details of LERP} We implement
LERP in PyTorch by following GraphHop. That is, we train two
independent LR classifiers with one- and two-hop neighborhood
aggregation. Thus, there are two LR classifiers trained in the label
embedding initialization stage and in the alternate optimization
process, respectively. For classifier training, we adopt the Adam
optimizer with a learning rate of $0.01$ and $5\times10^{-5}$ weight
decay for regularization.  Minibatch training is adopted for larger
label rates, while full batch training is used for smaller label
rates since there may not be enough labeled samples in a minibatch for
the latter. The number of training epochs is set to $1000$ with an early
stopping criterion to avoid overfitting. It is observed that the
above-mentioned hyperparameters have little impact on classifier
training in experiments; namely, the classifier converges efficiently
for a wide range of hyperparameters. 

We set the maximum number of alternate rounds in the alternate
optimization process to 100 for all datasets. Since this number is large
enough for LERP to converge as observed in the experiments. We
perform a grid search based on validation results in the hyperparameter
space of $T$, $\alpha$, $\beta$, and $max\_iter$. The temperature, $T$,
in the sharpening function is searched from grid $\{0.1, 0.5, 1, 10,
100\}$, the weighting parameter $\alpha$ in the classifier objective
function is searched from grid $\{0.1, 1, 10, 100\}$, the weighting
parameter $\beta$ in the label embedding iteration is searched from grid
$\{0.1, 0.5, 0.9\}$, and the number of iterations $max\_iter$ is
searched from grid $\{1, 5, 10\}$. These hyperparameters are tuned for
different label rates and datasets. 

%%%%%%%%%%%%%%%%%%%%%%%%%%%%%%%%%%%%%%%%%%%%%%%%%%%%%%%%%%%%%%
\begin{table*}[!htp]
\renewcommand{\arraystretch}{.9}
\caption{Test accuracy for the Cora dataset with extremely low label rates measured by ``mean
accuracy ($\%$) $\pm$ standard deviation". The highest mean accuracy is
in \textbf{bold} while the second and third ones are \underline{underlined}.}\label{tab:results_cora}
\centering
\resizebox{\textwidth}{!}{
\begin{tabular}{c|cccccc}
    \hline
    \multicolumn{7}{c}{\textbf{Cora}} \\
    \hline
    \# of labels per class & 1 & 2 & 4 & 8 & 16 & 20 \\
    \hline \hline
    \textbf{GCN} \cite{kipf2016semi} & $40.48\pm3.62$ & $49.70\pm1.56$ & $67.23\pm1.34$ & $73.88\pm0.75$ & $79.66\pm0.45$ & $81.76\pm0.25$ \\
    \textbf{GAT} \cite{velickovic2018graph} & $40.12\pm5.75$ & $50.69\pm2.08$ & $68.82\pm2.00$ & $75.08\pm0.81$ & $79.45\pm0.59$ & $81.59\pm0.44$ \\
%   \hline
    \textbf{LP} \cite{zhou2003learning} & $51.34\pm0.00$ & $54.19\pm0.00$ & $60.64\pm0.00$ & $67.57\pm0.00$ & $69.47\pm0.00$ & $71.03\pm0.00$ \\
    \textbf{APPNP} \cite{klicpera2018predict} & $60.70\pm1.26$ & \underline{$68.49\pm1.31$} & \underline{$75.12\pm0.93$} & \underline{$79.15\pm0.53$} & \underline{$81.17\pm0.40$} & \underline{${82.46\pm0.61}$}\\
    \textbf{C\&S} \cite{huang2020combining} & $37.46 \pm 14.90$ & $33.93\pm8.61$ & $48.93\pm9.94$ & $64.07\pm2.62$ & $73.63\pm1.14$ & $74.90\pm1.06$\\ 
    \textbf{GraphHop} \cite{xie2022graphhop} & $59.12\pm3.18$ & $59.21\pm2.66$ & \underline{$73.22\pm0.86$} & $75.48\pm0.90$ & $79.29\pm0.46$ & $81.05\pm0.39$ \\
%   \hline
    \textbf{Co-training GCN} \cite{li2018deeper} & $56.58\pm0.80$ & \underline{$66.92\pm0.73$} & $71.95\pm0.65$ & $75.56\pm0.93$ & $77.36\pm0.76$ & $80.17\pm0.97$\\
    \textbf{Self-training GCN} \cite{li2018deeper} & $39.71\pm3.42$ & $52.32\pm6.72$ & $65.82\pm4.85$ & $75.16\pm1.90$ & $78.14\pm0.92$ & $80.57\pm0.59$ \\
    \textbf{IGCN} \cite{li2019label} & \underline{$61.23\pm1.94$} & $63.75\pm2.59$ & $71.43\pm0.66$ & \underline{$78.46\pm0.57$} & $80.04\pm0.50$ & \underline{$82.51\pm0.41$} \\
    \textbf{GLP} \cite{li2019label} & $55.67\pm6.79$ & $57.71\pm3.99$ & $70.26\pm2.68$ & $76.78\pm1.17$ & \underline{$80.38\pm0.60$} & $82.17\pm0.71$ \\
    \textbf{CGPN} \cite{wan2021contrastive} & \underline{$70.56\pm0.00$} & $66.41\pm0.00$ & $72.71\pm0.00$ & $76.30\pm0.00$ & $75.91\pm0.00$ & $78.14\pm0.00$ \\
%   \hline  \hline
    \textbf{LERP} & $\mathbf{72.47\pm0.50}$ & $\mathbf{73.86\pm0.67}$ & $\mathbf{79.15\pm0.52}$ & $\mathbf{80.07\pm0.22}$ & $\mathbf{81.27\pm0.56}$ & $\mathbf{82.65\pm0.30}$\\
    \hline
\end{tabular}}
\end{table*}
%%%%%%%%%%%%%%%%%%%%%%%%%%%%%%%%%%%%%%%%%%%%%%%%%%%%%%%%%%%%%%

%%%%%%%%%%%%%%%%%%%%%%%%%%%%%%%%%%%%%%%%%%%%%%%%%%%%%%%%%%%%%%
\begin{table*}[!htp]
\renewcommand{\arraystretch}{.9}
\caption{Test accuracy for the CiteSeer dataset with extremely low label rates measured by 
``mean accuracy ($\%$) $\pm$ standard deviation". The highest mean accuracy is
in \textbf{bold} while the second and third ones are \underline{underlined}.}\label{tab:results_citeseer}
\centering
\resizebox{\textwidth}{!}{
\begin{tabular}{c|cccccc}
    \hline
    \multicolumn{7}{c}{\textbf{CiteSeer}} \\
    \hline
    \# of labels per class & 1 & 2 & 4 & 8 & 16 & 20 \\
    \hline \hline
    \textbf{GCN} \cite{kipf2016semi} & $26.04\pm1.65$ & $34.69\pm3.25$ & $52.02\pm0.75$ & $61.51\pm0.87$ & $68.03\pm0.36$ & $68.67\pm0.26$ \\
    \textbf{GAT} \cite{velickovic2018graph} & $26.83\pm2.32$ & $41.95\pm3.99$ & $52.43\pm0.97$ & $62.86\pm0.91$ & $68.23\pm0.46$ & $68.47\pm0.51$ \\
%   \hline
    \textbf{LP} \cite{zhou2003learning} & $20.10\pm0.00$ & $32.72\pm0.00$ & $33.25\pm0.00$ & $42.21\pm0.00$ & $46.58\pm0.00$ & $47.51\pm0.00$ \\
    \textbf{APPNP} \cite{klicpera2018predict} & $34.18\pm1.53$ & $47.04\pm2.64$ & \underline{$54.55\pm0.60$} & \underline{$65.40\pm0.47$} & \underline{$69.46\pm0.48$} & \underline{$70.32\pm0.72$} \\
    \textbf{C\&S} \cite{huang2020combining} & $25.34\pm4.88$ & $25.47\pm3.39$ & $37.55\pm3.19$ & $46.24\pm1.41$ & $55.68\pm1.47$ & $57.98\pm1.22$ \\
    \textbf{GraphHop} \cite{xie2022graphhop} & \underline{$48.40\pm3.08$} & \underline{$53.27\pm5.16$} & \underline{$54.34\pm1.66$} & $60.11\pm1.60$ & $64.99\pm1.26$ & $67.47\pm0.66$ \\
%   \hline
    \textbf{Co-training GCN} \cite{li2018deeper} & $28.24\pm0.27$ & $36.55\pm0.99$ & $33.77\pm3.62$ & $58.20\pm1.06$ & $64.10\pm1.69$ & $67.79\pm0.76$ \\
    \textbf{Self-training GCN} \cite{li2018deeper} & $30.45\pm5.76$ & $36.34\pm8.54$ & $43.59\pm5.91$ & $62.50\pm3.23$ & $68.41\pm0.84$ & $69.63\pm0.31$ \\
    \textbf{IGCN} \cite{li2019label} & $29.63\pm0.45$ & $45.19\pm1.18$ & $51.48\pm1.88$ & \underline{$64.49\pm1.03$} & \underline{$68.97\pm0.43$} & \underline{$69.80\pm0.28$} \\
    \textbf{GLP} \cite{li2019label} & $24.10\pm5.67$ & $40.00\pm5.45$ & $49.83\pm2.36$ & $63.54\pm1.16$ & $68.19\pm0.89$ & $69.10\pm0.37$ \\
    \textbf{CGPN} \cite{wan2021contrastive} & \underline{$51.93\pm0.00$} & $\mathbf{62.31\pm0.00}$ & $50.55\pm0.00$ & $59.63\pm0.00$ & $63.05\pm0.00$ & $62.91\pm0.00$\\
%   \hline \hline
    \textbf{LERP} & $\mathbf{53.20\pm0.85}$ & \underline{${59.66\pm0.44}$} & $\mathbf{61.04\pm1.24}$ & $\mathbf{66.39\pm0.44}$ & $\mathbf{69.62\pm0.32}$ & $\mathbf{70.77\pm0.47}$\\
    \hline
\end{tabular}}
\end{table*}
%%%%%%%%%%%%%%%%%%%%%%%%%%%%%%%%%%%%%%%%%%%%%%%%%%%%%%%%%%%%%%

%%%%%%%%%%%%%%%%%%%%%%%%%%%%%%%%%%%%%%%%%%%%%%%%%%%%%%%%%%%%%%
\begin{table*}[!htp]
\renewcommand{\arraystretch}{.9}
\caption{Test accuracy for the PubMed dataset with extremely low label rates measured by 
``mean accuracy ($\%$) $\pm$ standard deviation". The highest mean accuracy is
in \textbf{bold} while the second and third ones are \underline{underlined}.}\label{tab:results_pubmed}
\centering
\resizebox{\textwidth}{!}{
\begin{tabular}{c|cccccc}
    \hline
    \multicolumn{7}{c}{\textbf{PubMed}} \\
    \hline
    \# of labels per class & 1 & 2 & 4 & 8 & 16 & 20 \\
    \hline \hline
    \textbf{GCN} \cite{kipf2016semi} & $48.11\pm9.76$ & $65.01\pm2.05$ & \underline{$70.99\pm0.35$} & $70.57\pm0.79$ & $76.84\pm0.26$ & $77.38\pm0.20$ \\
    \textbf{GAT} \cite{velickovic2018graph} & $57.64\pm5.52$ & $70.04\pm1.14$ & $69.79\pm0.39$ & $71.66\pm0.30$ & $75.62\pm0.35$ & $76.46\pm0.20$ \\
%   \hline
    \textbf{LP} \cite{zhou2003learning} & $63.70\pm0.00$ & $67.16\pm0.00$ & $66.37\pm0.00$ & $65.89\pm0.00$ & $68.63\pm0.00$ & $70.55\pm0.00$ \\
    \textbf{APPNP} \cite{klicpera2018predict} & \underline{${71.05\pm0.37}$} & \underline{${71.37\pm0.35}$} & \underline{${71.05\pm0.19}$} & \underline{$72.73\pm1.22$} & $\mathbf{79.17\pm0.39}$ & \underline{$79.22\pm0.41$} \\
    \textbf{C\&S} \cite{huang2020combining} & $46.58\pm6.26$ & $57.41\pm8.33$ & $67.32\pm2.00$ & $71.08\pm0.99$ & $74.25\pm0.92$ & $74.51\pm0.85$ \\
    \textbf{GraphHop} \cite{xie2022graphhop} & $67.13\pm2.51$ & $68.82\pm0.80$ & $69.62\pm0.31$ & $71.21\pm0.58$ & $74.98\pm0.31$ & $76.05\pm0.32$ \\
%   \hline
    \textbf{Co-training GCN} \cite{li2018deeper} & $62.41\pm0.35$ & $68.70\pm0.31$ & $67.27\pm0.43$ & $69.66\pm0.17$ & $76.21\pm0.23$ & $76.93\pm0.20$ \\
    \textbf{Self-training GCN} \cite{li2018deeper} & $54.06\pm7.89$ & $70.50\pm2.83$ & $67.43\pm1.39$ & $69.40\pm1.03$ & $73.44\pm1.92$ & $76.96\pm0.85$ \\
    \textbf{IGCN} \cite{li2019label} & \underline{$70.17\pm0.11$} & \underline{$71.62\pm0.11$} & $70.93\pm0.08$ & $\mathbf{73.45\pm0.41}$ & \underline{$78.96\pm0.28$} & \underline{$79.53\pm0.15$} \\
    \textbf{GLP} \cite{li2019label} & $70.06\pm0.32$ & $71.30\pm0.41$ & $70.66\pm0.68$ & \underline{$73.39\pm0.33$} & \underline{$77.92\pm0.32$} & $79.13\pm0.31$ \\
    \textbf{CGPN} \cite{wan2021contrastive} & $69.22\pm0.00$ & $69.48\pm0.00$ & $68.38\pm0.00$ & $68.97\pm0.00$ & $68.87\pm0.00$ & $69.93\pm0.00$  \\
%   \hline \hline
    \textbf{LERP} & $\mathbf{72.67\pm1.79}$ & $\mathbf{72.00\pm0.36}$ & $\mathbf{71.11\pm0.19}$ & $72.59\pm0.07$ & $77.20\pm1.45$ & $\mathbf{79.71\pm0.33}$ \\
    \hline
\end{tabular}}
\end{table*}
%%%%%%%%%%%%%%%%%%%%%%%%%%%%%%%%%%%%%%%%%%%%%%%%%%%%%%%%%%%%%%

%%%%%%%%%%%%%%%%%%%%%%%%%%%%%%%%%%%%%%%%%%%%%%%%%%%%%%%%%%%%%%
\begin{table*}[!ht]
\renewcommand{\arraystretch}{.9}
\caption{Test accuracy for the Amazon Photo dataset with extremely low label rates measured by 
``mean accuracy ($\%$) $\pm$ standard deviation". The highest mean accuracy is
in \textbf{bold} while the second and third ones are \underline{underlined}.}\label{tab:results_photo}
\centering
\resizebox{\textwidth}{!}{
\begin{tabular}{c|cccccc}
    \hline
    \multicolumn{7}{c}{\textbf{Amazon Photo}} \\
    \hline
    \# of labels per class & 1 & 2 & 4 & 8 & 16 & 20 \\
    \hline \hline
    \textbf{GCN} \cite{kipf2016semi} & $35.61\pm3.37$ & $39.52\pm10.35$ & $69.55\pm1.37$ & $72.53\pm0.90$ & $75.76\pm1.06$ & $78.21\pm0.55$ \\
    \textbf{GAT} \cite{velickovic2018graph} & $46.24\pm0.00$ & $67.66\pm2.52$ & \underline{$78.78\pm4.24$} & $83.30\pm1.86$ & $83.54\pm1.47$ & $84.78\pm1.11$ \\
%   \hline
    \textbf{LP} \cite{zhou2003learning} & \underline{$61.09\pm0.00$} & \underline{$72.96\pm0.00$} & $67.79\pm0.00$ & $76.69\pm0.00$ & $81.32\pm0.00$ & $82.62\pm0.00$ \\
    \textbf{APPNP} \cite{klicpera2018predict} & $36.72\pm29.73$ & $33.15\pm32.40$ & $50.90\pm35.43$ & \underline{$86.63\pm0.56$} & \underline{$88.19\pm0.57$} & \underline{$87.50\pm0.60$} \\
    \textbf{C\&S} \cite{huang2020combining} & $33.01\pm18.02$ & $36.47\pm5.19$ & $64.45\pm4.27$ & $78.69\pm2.44$ & $84.67\pm1.18$ & $85.39\pm1.37$ \\ 
    \textbf{GraphHop} \cite{xie2022graphhop} & $58.76\pm4.12$ & \underline{$70.86\pm7.51$} & $78.30\pm2.78$ & \underline{$83.67\pm1.24$} & \underline{$87.16\pm1.55$} & \underline{$88.88\pm0.97$} \\
%   \hline
    \textbf{Co-training GCN} \cite{li2018deeper} & $45.97\pm6.00$ & $56.84\pm5.29$ & $70.82\pm3.44$ & $75.94\pm2.92$ & $79.49\pm1.11$ & $81.45\pm1.05$\\
    \textbf{Self-training GCN} \cite{li2018deeper} & $20.21\pm11.28$ & $28.65\pm8.98$ & $65.71\pm4.80$ & $72.69\pm4.61$ & $80.56\pm2.14$ & $83.09\pm0.87$ \\
    \textbf{IGCN} \cite{li2019label} & $29.57\pm7.03$ & $28.70\pm4.06$ & $56.51\pm1.41$ & $67.12\pm3.24$ & $72.77\pm4.10$ & $79.61\pm1.14$ \\
    \textbf{GLP} \cite{li2019label} & $9.12\pm0.00$ & $9.25\pm0.00$ & $26.59\pm4.20$ & $35.58\pm0.26$ & $40.49\pm5.74$ & $55.30\pm6.56$ \\
    \textbf{CGPN} \cite{wan2021contrastive} & \underline{$62.22\pm0.00$} & $64.76\pm0.00$ & \underline{$78.35\pm0.00$} & $80.03\pm0.00$ & $78.14\pm0.00$ & $82.16\pm0.00$ \\
%   \hline \hline
    \textbf{LERP} & $\mathbf{67.41\pm2.00}$ & $\mathbf{83.10\pm1.58}$ & $\mathbf{86.10\pm0.75}$ & $\mathbf{89.26\pm0.36}$ & $\mathbf{91.83\pm0.33}$ & $\mathbf{92.06\pm0.29}$ \\
    \hline
\end{tabular}}
\end{table*}
%%%%%%%%%%%%%%%%%%%%%%%%%%%%%%%%%%%%%%%%%%%%%%%%%%%%%%%%%%%%%%

%%%%%%%%%%%%%%%%%%%%%%%%%%%%%%%%%%%%%%%%%%%%%%%%%%%%%%%%%%%%%%
\begin{table*}[!ht]
\renewcommand{\arraystretch}{.9}
\caption{Test accuracy for the Coauthor CS dataset with extremely low
label rates measured by ``mean accuracy ($\%$) $\pm$ standard
deviation". The highest mean accuracy is in \textbf{bold} while the second
and third ones are \underline{underlined}.}\label{tab:results_cs}
\centering
\resizebox{\textwidth}{!}{
\begin{tabular}{c|cccccc}
    \hline
    \multicolumn{7}{c}{\textbf{Coauthor CS}} \\
    \hline
    \# of labels per class & 1 & 2 & 4 & 8 & 16 & 20 \\
    \hline \hline
    \textbf{GCN} \cite{kipf2016semi} & $64.42\pm2.14$ & $74.07\pm1.09$ & $82.62\pm0.74$ & $87.88\pm0.55$ & \underline{$89.86\pm0.15$} & \underline{$90.19\pm0.17$} \\
    \textbf{GAT} \cite{velickovic2018graph} & \underline{$72.81\pm2.01$} & $80.76\pm1.25$ & $85.37\pm0.94$ & $88.06\pm0.61$ & $89.83\pm0.16$ & $89.82\pm0.17$ \\
%   \hline
    \textbf{LP} \cite{zhou2003learning} & $52.63\pm0.00$ & $59.34\pm0.00$ & $61.77\pm0.00$ & $68.60\pm0.00$ & $73.50\pm0.00$ & $74.65\pm0.00$ \\
    \textbf{APPNP} \cite{klicpera2018predict} & $71.06\pm17.48$ & \underline{$86.18\pm0.56$} & ${86.57\pm0.58}$ & \underline{$89.66\pm0.30$} & \underline{$90.65\pm0.21$} & \underline{$90.55\pm0.17$} \\
    \textbf{C\&S} \cite{huang2020combining} & $35.63\pm13.20$ & $67.11\pm3.50$ & $78.57\pm1.22$ & $83.92\pm1.15$ & $87.09\pm0.95$ & $87.97\pm0.56$ \\
    \textbf{GraphHop} \cite{xie2022graphhop} & $65.03\pm0.01$ & $77.59\pm3.17$ & $83.79\pm1.13$ & $86.69\pm0.62$ & $89.47\pm0.38$ & $89.84\pm0.00$ \\
%   \hline
    \textbf{Co-training GCN} \cite{li2018deeper} & \underline{$75.78\pm1.00$} & $\mathbf{86.94\pm0.70}$ & \underline{$87.40\pm0.78$} & $88.81\pm0.32$ & $89.22\pm0.42$ & $89.01\pm0.55$ \\
    \textbf{Self-training GCN} \cite{li2018deeper} & $69.69\pm3.27$ & $82.79\pm4.10$ & \underline{$87.62\pm1.49$} & \underline{$88.82\pm1.00$} & $89.53\pm0.54$ & $89.07\pm0.70$ \\
    \textbf{IGCN} \cite{li2019label} & $62.16\pm2.81$ & $59.56\pm2.91$ & $65.82\pm4.86$ & $86.57\pm0.78$ & $87.76\pm0.74$ & $88.10\pm0.52$ \\
    \textbf{GLP} \cite{li2019label} & $43.56\pm7.06$ & $50.74\pm7.55$ & $46.61\pm9.85$ & $76.61\pm3.39$ & $81.75\pm2.81$ & $82.43\pm3.31$ \\
    \textbf{CGPN} \cite{wan2021contrastive} & $67.66\pm0.00$ & $64.49\pm0.00$ & $71.00\pm0.00$ & $77.09\pm0.00$ & $78.75\pm0.00$ & $79.71\pm0.00$ \\
%   \hline \hline
    \textbf{LERP} & $\mathbf{82.46\pm1.28}$ & \underline{$86.37\pm0.37$} & $\mathbf{88.45\pm0.35}$ & $\mathbf{89.87\pm0.48}$ & $\mathbf{90.69\pm0.13}$ & $\mathbf{90.87\pm0.06}$ \\
    \hline
\end{tabular}}
\end{table*}
%%%%%%%%%%%%%%%%%%%%%%%%%%%%%%%%%%%%%%%%%%%%%%%%%%%%%%%%%%%%%%

\subsection{Performance Evaluation}\label{subsec:evaluation}

The performance results on five graph datasets are summarized in Tables
\ref{tab:results_cora}, \ref{tab:results_citeseer},
\ref{tab:results_pubmed}, \ref{tab:results_photo}, and
\ref{tab:results_cs}, respectively. Each column shows the classification
accuracy (\%) of LERP and benchmarking methods on test data under
a specific label rate. Overall, LERP has the top performance among
all comparators in most cases.  In particular, for cases with extremely
limited labels, LERP outperforms other benchmarking methods by a
large margin. This is because graph convolutional networks are difficult
to train with a small number of labels. The lack of supervision prevents
them from learning the transformation from the input feature space to
the label embedding space with nonlinear activation at each layer.
Instead, LERP applies regularization to the label embedding space.
It is more effective since it relieves the burden of learning the
transformation. A small number of labeled nodes also restricts the
efficacy of message passing on graphs of GCN-based methods. In general,
only two convolutional layers are adopted by GCN-based methods, which
means that only messages in the two-hop neighborhood of each labeled
node can be supervised.  However, the two-hop neighborhood of a limited
number of labeled nodes cannot cover the whole network effectively. 
Consequently, a large number of nodes do not have supervised training from
labels, resulting in an inferior performance of all GCN-based methods.
Some label-efficient GCN-based methods try to alleviate this problem by
exploiting pseudo-labels as supervision (e.g., self-training GCN) or
improving message passing capability (e.g., IGCN and CGPN).  However,
they are still handicapped by deficient message passing in the graph
convolutional layers. 

Some propagation-based methods (e.g., LERP, APPNP, and GraphHop)
achieve better performance in most datasets with low label rates.  This
is because the messages from labeled nodes can pass a longer distance on
graphs through an iterative process. Yet, although LP and C\&S are also
propagation-based, their performance is poorer since LP fails to encode
the rich node attribute information in model learning while C\&S was
originally designed for supervised learning and lack of labeled samples
degrades its performance significantly. 

\subsection{Convergence Analysis}\label{subsec:convergence}

\begin{figure*}[!t]
\centering
\subfloat[]{\includegraphics[width=0.2\textwidth]{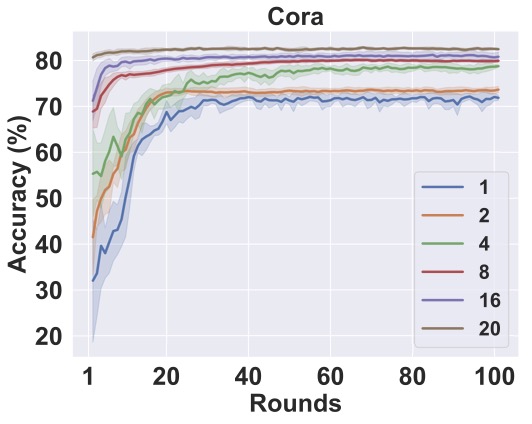}%
\label{fig:accuracy_round_cora}}
\hfil
\subfloat[]{\includegraphics[width=0.2\textwidth]{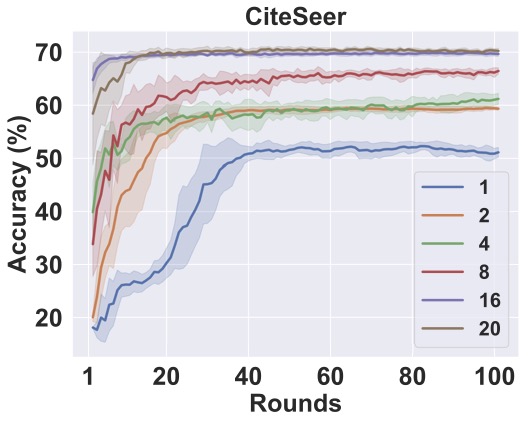}%
\label{fig:accuracy_round_citeseer}}
\hfil
\subfloat[]{\includegraphics[width=0.2\textwidth]{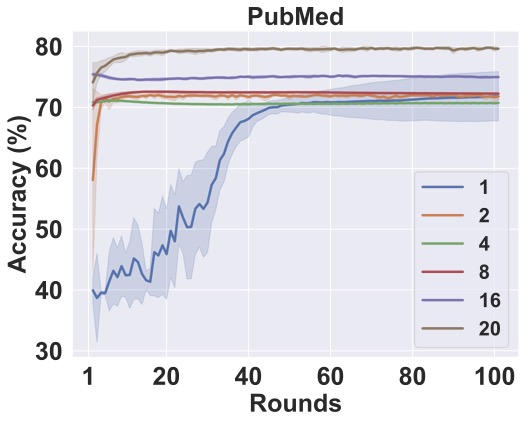}%
\label{fig:accuracy_round_pubmed}}
\hfil
\subfloat[]{\includegraphics[width=0.2\textwidth]{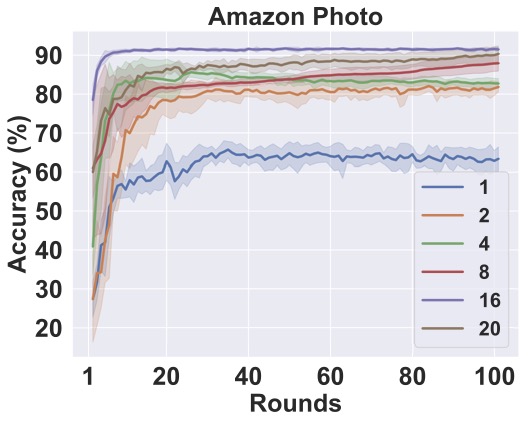}%
\label{fig:accuracy_round_photo}}
\hfil
\subfloat[]{\includegraphics[width=0.2\textwidth]{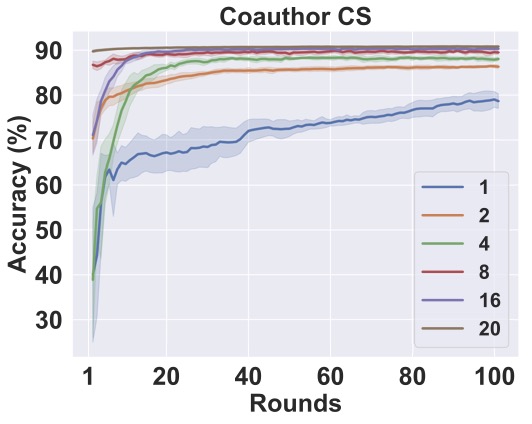}%
\label{fig:accuracy_round_cs}}
\caption{Convergence results of the label embeddings for the five
benchmarking datasets: (a) Cora, (b) CiteSeer, (c) PubMed, (d) Amazon
Photo, and (e) Coauthor CS.  The $x$-axis is the number of alternate
optimization rounds and the $y$-axis is the test accuracy ($\%$).
Different curves show the mean accuracy values under different label rates
and the shaded areas represent the standard deviation. }
\label{fig:accuracy_round}
\end{figure*}

\begin{figure*}[!t]
\centering
\subfloat[]{\includegraphics[width=0.2\textwidth]{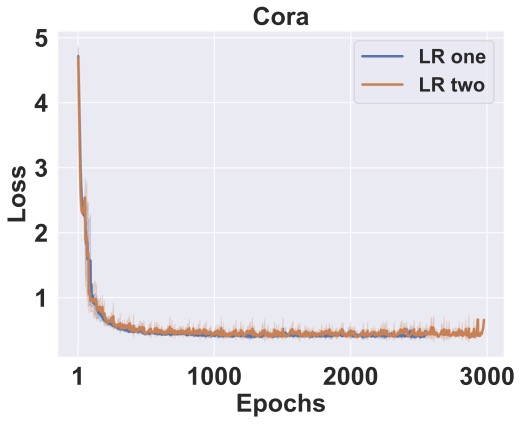}%
\label{fig:loss_epoch_cora}}
\hfil
\subfloat[]{\includegraphics[width=0.2\textwidth]{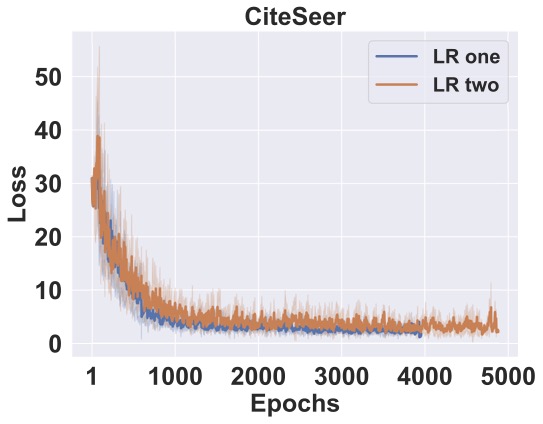}%
\label{fig:loss_epoch_citeseer}}
\hfil
\subfloat[]{\includegraphics[width=0.2\textwidth]{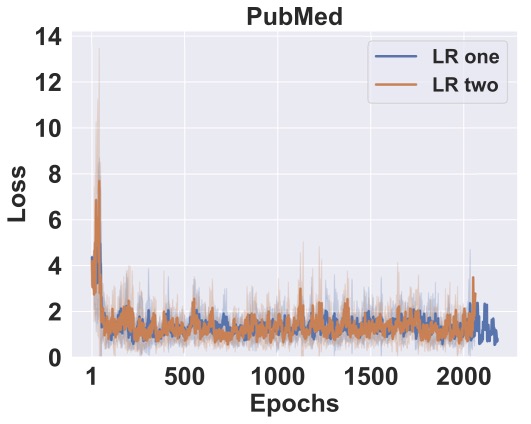}%
\label{fig:loss_epoch_pubmed}}
\hfil
\subfloat[]{\includegraphics[width=0.2\textwidth]{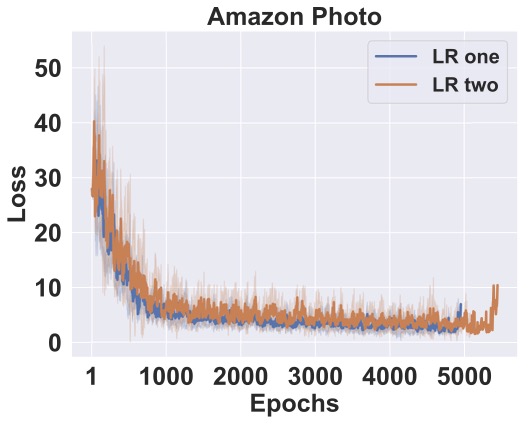}%
\label{fig:loss_epoch_photo}}
\hfil
\subfloat[]{\includegraphics[width=0.2\textwidth]{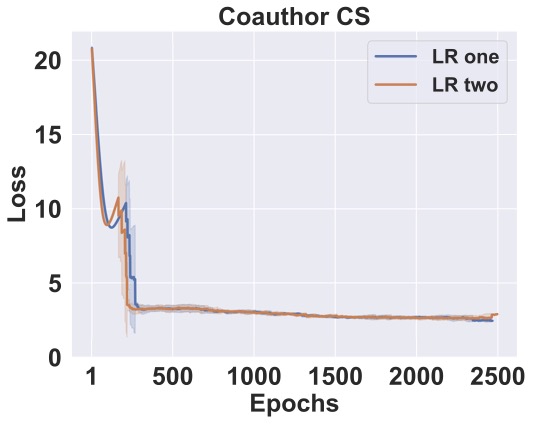}%
\label{fig:loss_epoch_cs}}
\caption{Convergence results of the LR classifiers for the five benchmarking
datasets: (a) Cora, (b) CiteSeer, (c) PubMed, (d) Amazon Photo, and (e)
Coauthor CS. The $x$-axis is the number of training epochs and the $y$-axis 
is the training loss. The label rate is 20 labels per class.}
\label{fig:loss_epoch}
\end{figure*}

Since LERP provides an alternate optimization solution to the problem
(\ref{equ:graphhop_variational}), we show the convergence of label
embeddings $\mathbf{F}$ and parameters of LR classifiers $\mathbf{W}$ in
Figs.  \ref{fig:accuracy_round} and \ref{fig:loss_epoch}, respectively,
to demonstrate its convergence behavior.  Fig.  \ref{fig:accuracy_round}
shows the accuracy change along with the number of alternate rounds of
the optimization process. We see that label embeddings converge in all
five benchmarking datasets under different label rates. A smaller label
rate requires a larger number of alternate rounds. Since classifiers are
trained on reliable nodes as defined in Eq.
(\ref{equ:graphhop_plus_classifier_loss}), it needs more alternate
rounds to train on the entire node set and propagate label embeddings to
the whole graph. Furthermore, the convergence speed is relatively fast.
The number of alternate rounds is usually less than 20 in most
investigated cases, leading to significant training efficiency as
discussed later. Fig.  \ref{fig:loss_epoch} depicts the cross-entropy
loss curves as a function of the training epoch for the two LR
classifiers at a label rate of 20 labels per class. The number of epochs
is counted throughout the whole optimization process. As shown in the
figure, the curves converge rapidly in several epochs. 

%%%%%%%%%%%%%%%%%%%%%%%%%%%%%%%%%%%%%%%%%%%%%%%%%%%%%
%%%%%%%%%%%%%%%%%%%%%%%%%%%%%%%%%%%%%%%%%%%%%%%%%%%%%

%%%%%%%%%%%%%%%%%%%%%%%%%%%%%%%%%%%%%%%%%%%%%%%%%%%%%
%%%%%%%%%%%%%%%%%%%%%%%%%%%%%%%%%%%%%%%%%%%%%%%%%%%%%

\subsection{Complexity and Memory Requirements}\label{subsec:complexity}

\begin{figure*}[!t]
\centering
\subfloat[]{\includegraphics[width=0.2\textwidth]{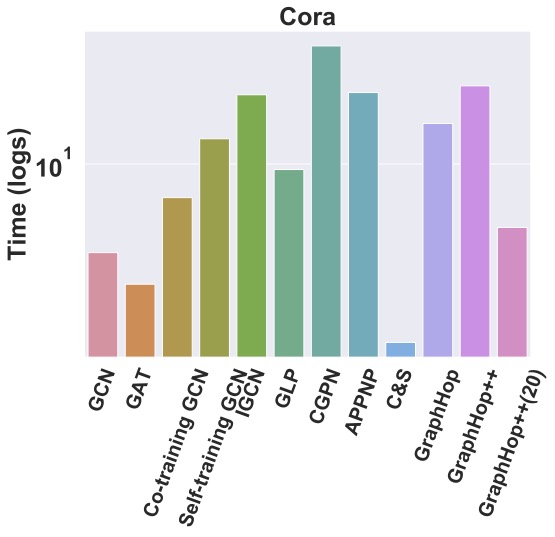}%
\label{fig:time_model_cora}}
\hfil
\subfloat[]{\includegraphics[width=0.2\textwidth]{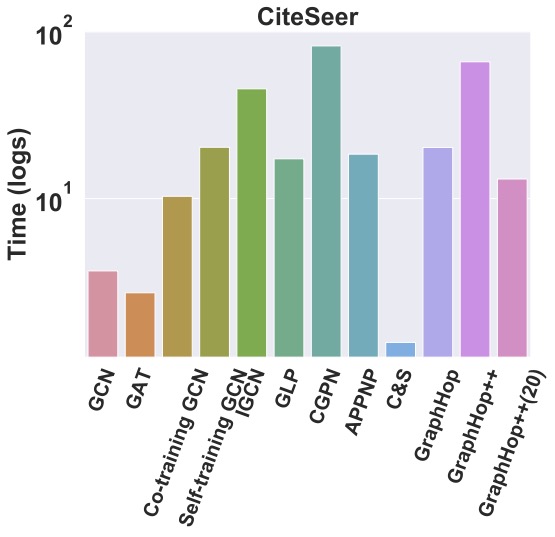}%
\label{fig:time_model_citeseer}}
\hfil
\subfloat[]{\includegraphics[width=0.2\textwidth]{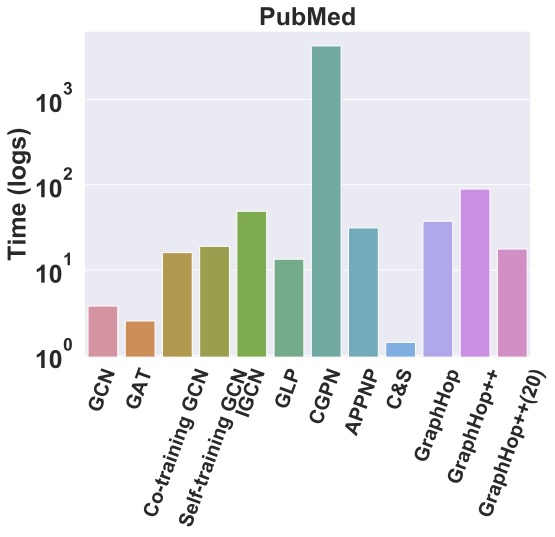}%
\label{fig:time_model_pubmed}}
\hfil
\subfloat[]{\includegraphics[width=0.2\textwidth]{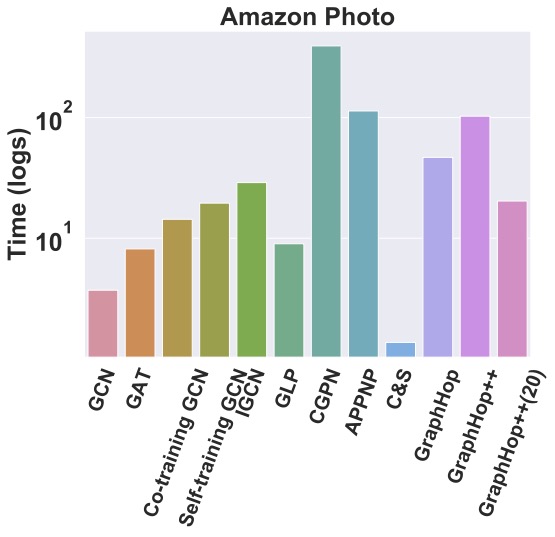}%
\label{fig:time_model_photo}}
\hfil
\subfloat[]{\includegraphics[width=0.2\textwidth]{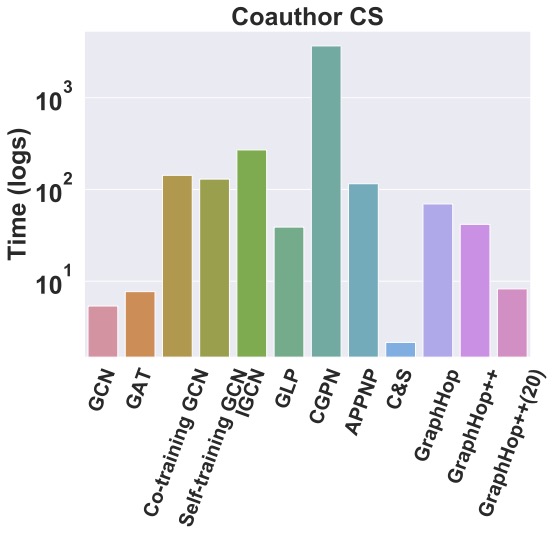}%
\label{fig:time_model_cs}}
\caption{Comparsion of computational efficiency of different methods
measured by log(second) for (a) Cora, (b) CiteSeer, (c) PubMed, (d) Amazon Photo
, and (e) Coauthor CS datasets, where the label rate is 20 labeled samples per
class and LERP(20) is the result of LERP with 20 alternate
optimization rounds.} \label{fig:time_model}
\end{figure*}

We compare the running time performance of LERP and it's several
benchmarking methods.  Fig.  \ref{fig:time_model} shows the results at a
label rate of 20 labeled samples per class with respect to the five
datasets. For LERP, we report results with 100 rounds and 20
rounds, respectively, since it can converge within 20 rounds as
indicated in Fig.  \ref{fig:accuracy_round}. We see from Fig.
\ref{fig:time_model} that LERP(20) can achieve an average running
time. The most time-consuming part of LERP is the training of the
LR classifiers, which demands multiple loops of all data samples.
However, we find that incorporating classifiers is essential to the
superior performance of LERP under limited label rates as discussed later. In other
words, LERP trades some time for effectiveness in the case of
extremely low label rates. Among all datasets, C\&S achieves the
lowest running time due to its simple design as LP. 

\begin{figure*}[!t]
\centering
\subfloat[]{\includegraphics[width=0.2\textwidth]{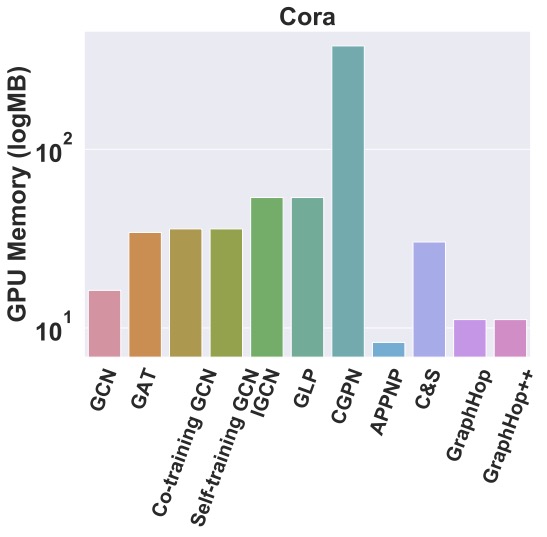}%
\label{fig:memory_model_cora}}
\hfil
\subfloat[]{\includegraphics[width=0.2\textwidth]{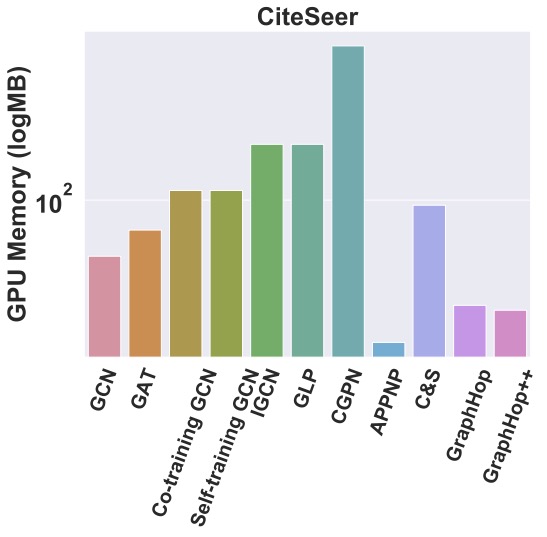}%
\label{fig:memory_model_citeseer}}
\hfil
\subfloat[]{\includegraphics[width=0.2\textwidth]{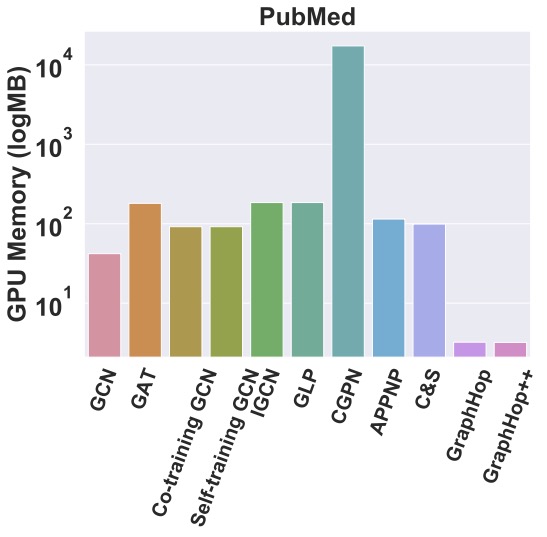}%
\label{fig:memory_model_pubmed}}
\hfil
\subfloat[]{\includegraphics[width=0.2\textwidth]{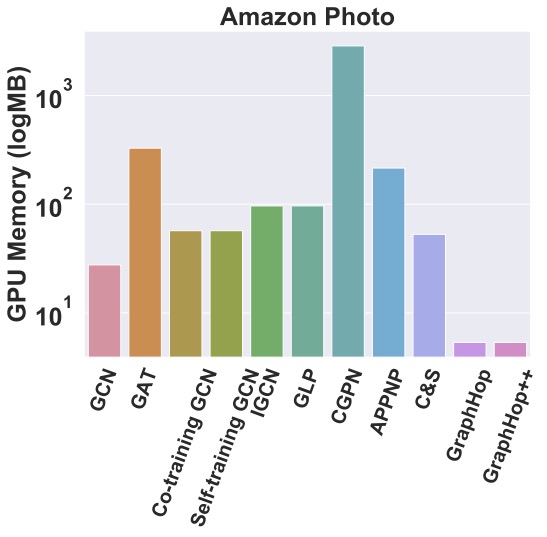}%
\label{fig:memory_model_photo}}
\hfil
\subfloat[]{\includegraphics[width=0.2\textwidth]{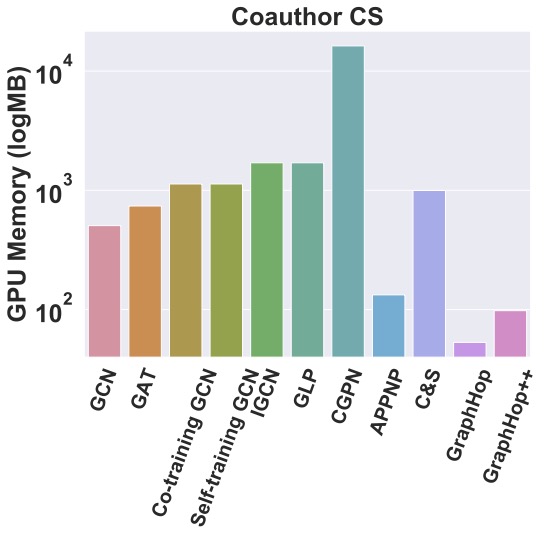}%
\label{fig:memory_model_cs}}
\caption{Comparsion of of GPU memory usages of different methods
measured by log(Mega Bytes) for (a) Cora, (b) CiteSeer, (c) PubMed, (d) Amazon
Photo, and (e) Coauthor CS datasets, where the label rate is 20 labeled
samples per class.} \label{fig:memory_model}
\end{figure*}

Next, we compare the GPU memory usage of LERP and its several
benchmarking methods\footnote{It is measured by
\textsf{{torch.cuda.max\_memory\_allocated()}} for PyTorch and
\textsf{{tf.contrib.memory\_stats.{MaxBytesInUse}()}} for TensorFlow.}.
The results are shown in Fig. \ref{fig:memory_model}, where the values
of LERP are taken from the same experiment of running time
comparison as given in Fig.  \ref{fig:time_model}. Generally, GraphHop
and LERP achieve the lowest GPU memory usage among all
benchmarking methods against the five datasets.  The reason is that
GraphHop and LERP allow minibatch training. The only parameters to
be stored in the GPU are classifier parameters and one minibatch of
data. Instead, the GCN-based methods cannot simply conduct minibatch
training. Since embeddings from different layers need to be stored for
backpropagation, the GPU memory consumption increases significantly. 

%%%%%%%%%%%%%%%%%%%%%%%%%%%%%%%%%%%%%%%%%%%%%%%%%%%%%
%%%%%%%%%%%%%%%%%%%%%%%%%%%%%%%%%%%%%%%%%%%%%%%%%%%%%

%%%%%%%%%%%%%%%%%%%%%%%%%%%%%%%%%%%%%%%%%%%%%%%%%%%%%
%%%%%%%%%%%%%%%%%%%%%%%%%%%%%%%%%%%%%%%%%%%%%%%%%%%%%

%%%%%%%%%%%%%%%%%%%%%%%%%%%%%%%%%%%%%%%%%%%%%%%%%%%%%
\begin{figure*}[!t]
\centering
\subfloat[]{\includegraphics[width=0.29\textwidth]{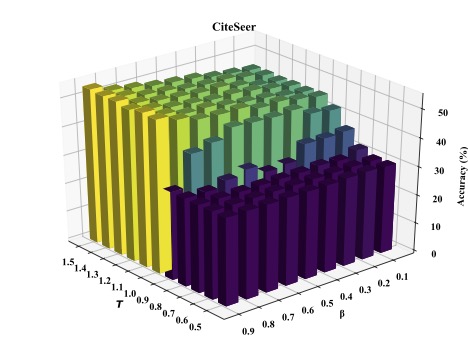}%
\label{fig:param_sens_citeseer_alpha_fix}}
\hfil
\subfloat[]{\includegraphics[width=0.29\textwidth]{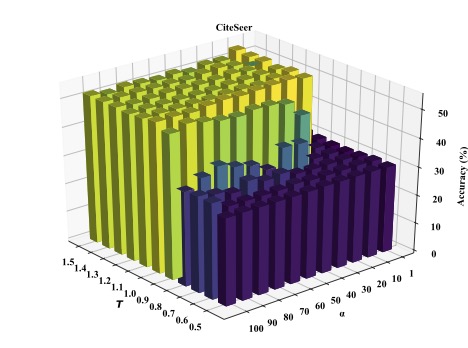}%
\label{fig:param_sens_citeseer_beta_fix}}
\hfil
\subfloat[]{\includegraphics[width=0.29\textwidth]{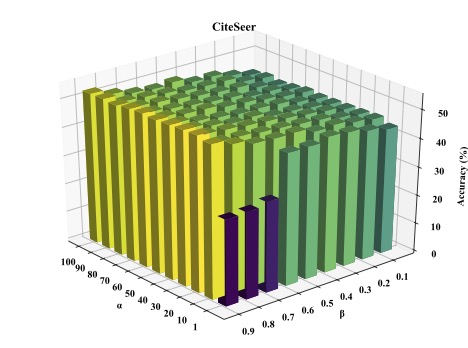}%
\label{fig:param_sens_citeseer_temperature_fix}}
\caption{Performance of LERP with various hyperparameter settings
for the CiteSeer dataset, where the $z$-axis is the accuracy result and
the label rate is one label per class: (a) performance of $T$ and
$\beta$ with fixed $\alpha$, (b) performance of $T$ and $\alpha$ with
fixed $\beta$, (c) performance of $\alpha$ and $\beta$ with fixed $T$.}
\label{fig:param_sens_citeseer}
\end{figure*}
%%%%%%%%%%%%%%%%%%%%%%%%%%%%%%%%%%%%%%%%%%%%%%%%%%%%%

%%%%%%%%%%%%%%%%%%%%%%%%%%%%%%%%%%%%%%%%%%%%%%%%%%%%%
\begin{figure*}[!t]
\centering
\subfloat[]{\includegraphics[width=0.2\textwidth]{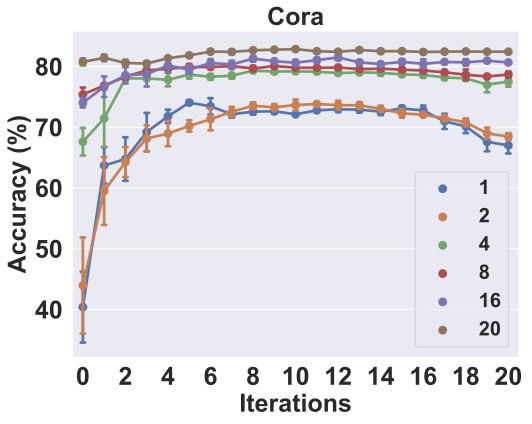}%
\label{fig:accuracy_iteration_cora}}
\hfil
\subfloat[]{\includegraphics[width=0.2\textwidth]{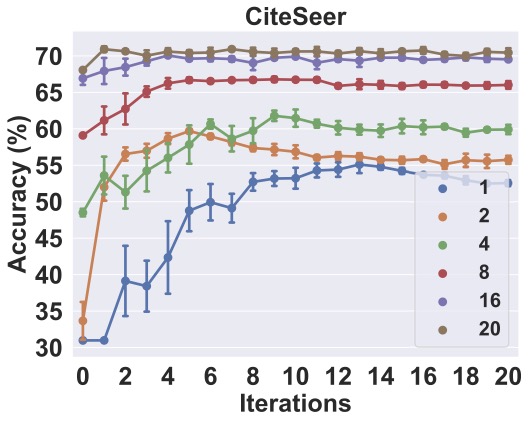}%
\label{fig:accuracy_iteration_citeseer}}
\hfil
\subfloat[]{\includegraphics[width=0.2\textwidth]{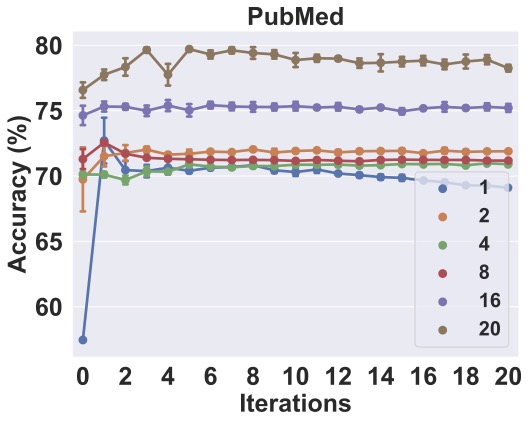}%
\label{fig:accuracy_iteration_pubmed}}
\hfil
\subfloat[]{\includegraphics[width=0.2\textwidth]{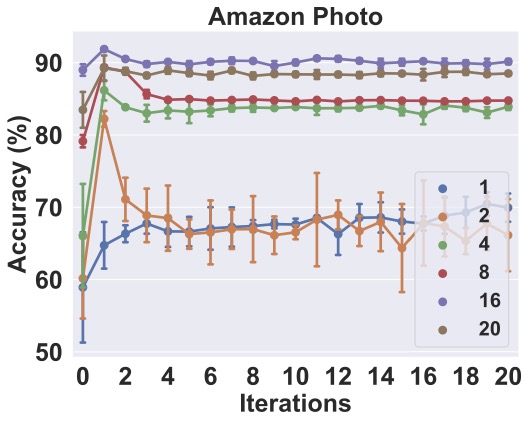}%
\label{fig:accuracy_iteration_photo}}
\hfil
\subfloat[]{\includegraphics[width=0.2\textwidth]{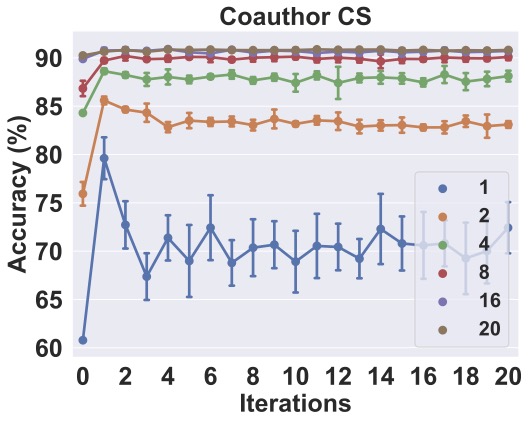}%
\label{fig:accuracy_iteration_cs}}
\caption{Performance of LERP with an increasing number of
iterations for five datasets: (a) Cora, (b) CiteSeer,
(c) PubMed, (d) Amazon Photo, and (e) Coauthor CS,
where the $x$-axis is the number of alternate optimization rounds and
the $y$-axis is the test accuracy ($\%$). The averaged accuracy under
one specific number of iterations is represented as dots and the
standard deviation as vertical bars. Different lines denote results
under different label rates.}
\label{fig:accuracy_iteration}
\end{figure*}
%%%%%%%%%%%%%%%%%%%%%%%%%%%%%%%%%%%%%%%%%%%%%%%%%%%%%

%%%%%%%%%%%%%%%%%%%%%%%%%%%%%%%%%%%%%%%%%%%%%%%%%%%%%
\begin{figure*}[!t]
\centering
\subfloat[]{\includegraphics[width=0.2\textwidth]{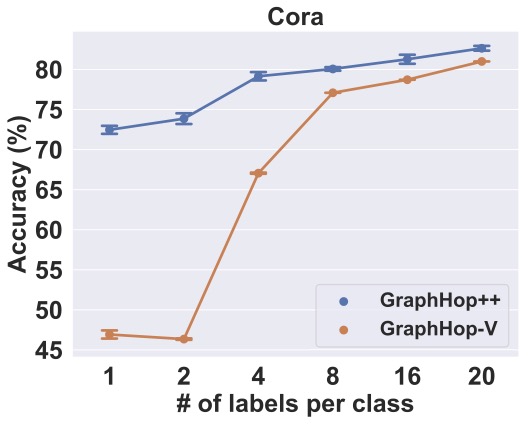}%
\label{fig:accuracy_num_labels_cora}}
\hfil
\subfloat[]{\includegraphics[width=0.2\textwidth]{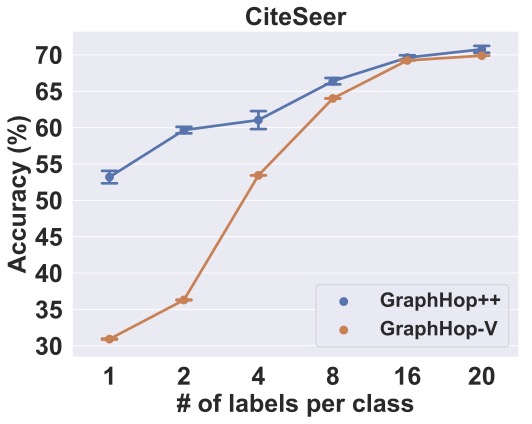}%
\label{fig:accuracy_num_labels_citeseer}}
\hfil
\subfloat[]{\includegraphics[width=0.2\textwidth]{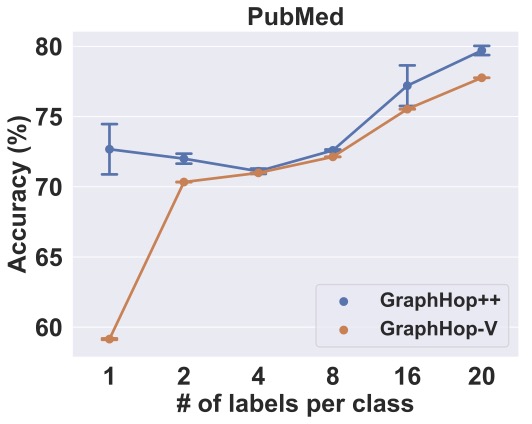}%
\label{fig:accuracy_num_labels_pubmed}}
\hfil
\subfloat[]{\includegraphics[width=0.2\textwidth]{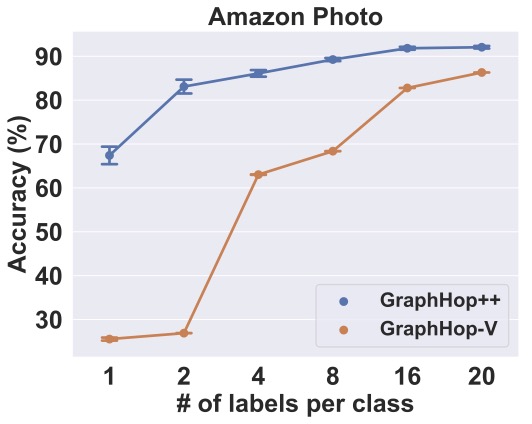}%
\label{fig:accuracy_num_labels_photo}}
\hfil
\subfloat[]{\includegraphics[width=0.2\textwidth]{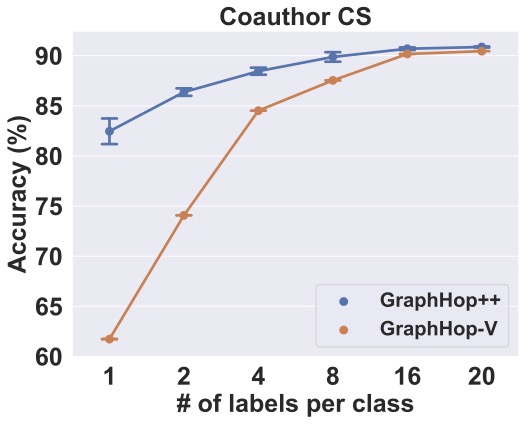}%
\label{fig:accuracy_num_labels_cs}}
\caption{Performance comparison of LERP and LERP-V under different label
rates: (a) Cora, (b) CiteSeer, (c) PubMed, (d) Amazon Photo, and (e) Coauthor CS datasets.} 
\label{fig:accuracy_num_labels}
\end{figure*}
%%%%%%%%%%%%%%%%%%%%%%%%%%%%%%%%%%%%%%%%%%%%%%%%%%%%%

\subsubsection{Parameter Sensitivity}

The effects of three hyperparameters, i.e., $T$, $\alpha$ and $\beta$,
of LERP on the performance are analyzed here. Due to page limitations,
we focus on a representative case in our discussion below, namely, the
CiteSeer dataset with a label rate of one-labeled sample per class. To
study the effect of three hyperparameters, we fix one and change the
other two using grid search. For example, we fix $T$ and adjust $\alpha$
and $\beta$. The sweeping ranges are around the adopted tuning ranges.
For each setting, results are averaged over ten runs. They are depicted
in Fig. \ref{fig:param_sens_citeseer}. 

We have the following observations.  First, hyperparameter $T$ has the
largest influence on the performance, and a small value will degrade the
accuracy result.  However, once $T>1$, the performance stays the same.
Note that $T=1$ means that there is no sharpening operation on the label
distribution in Eq.  (\ref{equ:label_update_loss_2}). In practice, we
can eliminate hyperparameter $T$ by removing the sharpening operation.
Second, $\alpha$ and $\beta$ have less influence on the performance.  By
increasing $\alpha$ and $\beta$ slightly, the performance improves as
shown in Fig. \ref{fig:param_sens_citeseer}. We need to tune these two
hyperparameters so as to achieve the best performance. 

\subsection{Ablation Study}\label{subsec:ablation}

\subsubsection{Enhancement via Label Embeddings Update} As compared to
GraphHop, LERP has an enhancement by conducting several iterations
of label embeddings in Eq.  (\ref{equ:graphhop_plus_iteration}). That
results in a better approximation to the optimum of Eq.
(\ref{equ:graphhop_optimize_f_fixed}). This is experimentally verified
below. We measure the accuracy under the same experimental settings but
vary the number of iterations, $max\_iter$, from zero to 20.  The
results against five datasets are shown in Fig.
\ref{fig:accuracy_iteration}. It is worthwhile to point out that
LERP degenerates to GraphHop with zero iteration, i.e., only
performing Eq.  (\ref{equ:graphhop_plus_iteration_init}).  We see from
the figure that increasing the number of iterations improves the
performance overall. This is especially true when the label rate is very
low. Fewer labeled nodes demand a larger number of iterations for better
approximation. Furthermore, we observe that the performance improvement
saturates within ten iterations.  Since several iterations can achieve
sufficient approximation with high accuracy, there is no need to reach
optimality in label embedding optimization.  It saves a considerable
amount of running time. 

\subsubsection{Removing LR Classifier Training} LERP is an
alternate approximate optimization of the variational problem given in
Eq.  (\ref{equ:graphhop_variational}). If we set hyperparameter
$\mathbf{U}_\alpha=\mathbf{0}$, the alternate optimization process
reduces to label embeddings optimization without the LR classifier
training.  Note that this is still different from the regularization
framework of the classical LP algorithm given in Eq.
(\ref{equ:regularization}), where the ground-truth label, $\mathbf{Y}$,
is changed to the initial label embeddings $\mathbf{F}_{init}$ as
supervision. This model without LR classifier training is named
``LERP-V (LERP-Variant)". We conduct experiments of LERP-V
on the same five datasets and show the results in Fig.
\ref{fig:accuracy_num_labels}.  We see that LERP outperforms
LERP-V on all datasets under various label rates. The performance
gap is especially obvious when the number of labeled samples is very
low. This can be explained as follows. Recall that LR classifiers are
trained on labeled and reliable nodes and used to infer label embeddings
for all nodes in the next round as given in Eq.
(\ref{equ:graphhop_plus_iteration_init}).  The inference improves the
propagation of label embedding from labeled nodes to the entire graph.
This is especially important when label rates are low since the
performance of the original propagation as given in Eq. (\ref{equ:lp})
is poor under this situation \cite{calder2020poisson}. Another
interesting observation is that this simple variant can achieve
performance comparable with that of GCN under the standard setting of 20
labels per class. 

\subsection{Application to Object Recognition}\label{subsec:object_recognition}

\begin{figure}[!t]
\centering
\includegraphics[width=0.45\textwidth]{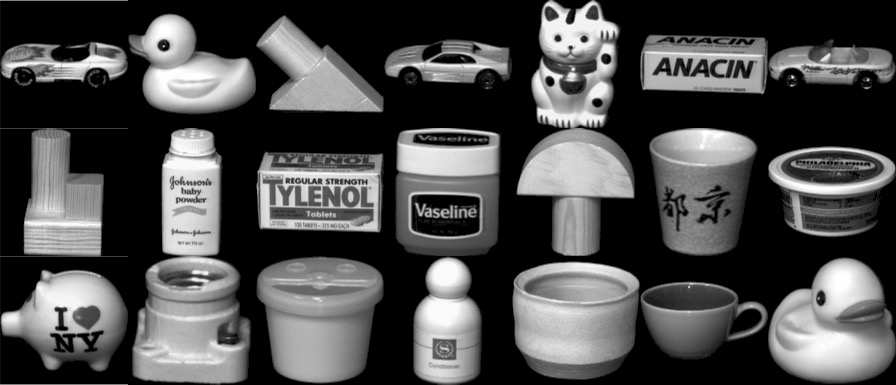}
\caption{Illustration of several exemplary images of the COIL20 dataset.}
\label{fig:sample_images}
\end{figure}

To show the effectiveness and other potential applications of LERP, we
apply it to an object recognition problem in this subsection. COIL20
\cite{nene1996columbia} is a popular dataset for object recognition.  It
contains 1,440 object images belonging to 20 classes, where each object
has 72 images shot from different angles. Some exemplary images are
given in Fig.  \ref{fig:sample_images}. The resolution of each image is
$32\times32$ and each pixel has 256 gray levels. As a result, each image
can be represented as a 1024-D vector. 

Each image corresponds to a node.  A k-nearest neighbor (kNN) graph with
$k=7$ is built based on the Euclidean distance of two images; namely,
edge weights are the Euclidean distance between two connected nodes.  We
conduct experiments on this constructed graph for LERP and other
benchmarking methods. We tune the hyperparameters of all benchmarks and
report the best accuracy performance. As for LERP, we set the
number of iterations to 50 and $\alpha=0.99$ for lower label rates. The
results under three different label rates are shown in Table
\ref{tab:results_coil20}. LERP achieves the best performance at
all three label rates.  The high performance of the LP method indicates
that the node attribute information (i.e., 1024 image pixels) does not
contribute much to label predictions, which is different from other
benchmarking datasets.  Instead, the manifold regularization of label
embeddings is the factor that is most relevant to high performance. This
explains the fact that propagation-based methods achieve better accuracy
than GCN-based methods at very low label rates. However, once there are
enough labeled samples (e.g., 20 labeled samples per class), GCN-based
or other propagation-based methods can achieve comparable performance or
outperform the LP method. In this case, there are sufficient labeled
samples for neural network training. They further exploit the node
attribute information in model learning. 

%%%%%%%%%%%%%%%%%%%%%%%%%%%%%%%%%%%%%%%%%%%%%%%%%%%%%%%%%%%%%%
\begin{table}[!t]
\renewcommand{\arraystretch}{.9}
\caption{Test accuracy with three label rates measured by ``mean accuracy ($\%$)
$\pm$ standard deviation", where the highest mean accuracy is marked in
\textbf{bold} and the second and third are \underline{underlined}.}\label{tab:results_coil20}
\centering
\resizebox{\hsize}{!}{
\begin{tabular}{c|ccc}
    \hline
    \multicolumn{4}{c}{\textbf{COIL20}} \\
    \hline
    \# of labels per class & 1 & 2 & 20 \\
    \hline \hline
    \textbf{GCN} \cite{kipf2016semi} & $61.05\pm1.85$ & $68.19\pm3.75$ & $88.72\pm1.64$ \\
    \textbf{GAT} \cite{velickovic2018graph} & $57.98\pm5.27$ & $65.32\pm3.36$ & $86.66\pm2.82$ \\
%   \hline
    \textbf{LP} \cite{zhou2003learning} & \underline{$83.48\pm0.00$} & \underline{$85.89\pm0.00$} & $88.89\pm0.00$ \\
    \textbf{APPNP} \cite{klicpera2018predict} & $78.11\pm0.23$ & \underline{$79.94\pm0.60$} & $90.67\pm0.45$ \\
    \textbf{C\&S} \cite{huang2020combining} & $67.85\pm4.66$ & $75.67\pm2.79$ & \underline{$92.50\pm0.68$} \\
    \textbf{GraphHop} \cite{xie2022graphhop} & \underline{$78.42\pm2.72$} & $78.99\pm2.95$ & \underline{$92.70\pm0.96$} \\
%   \hline
    \textbf{Co-training GCN} \cite{li2018deeper} & $60.07\pm3.04$ & $71.10\pm2.99$ & $91.22\pm1.12$ \\
    \textbf{Self-training GCN} \cite{li2018deeper} & $59.96\pm1.95$ & $70.32\pm3.44$ & $91.15\pm0.75$ \\
    \textbf{IGCN} \cite{li2019label} & $73.77\pm4.06$ & $75.70\pm2.46$ & $82.24\pm1.73$ \\
    \textbf{GLP} \cite{li2019label} &  $45.53\pm7.44$ & $39.79\pm4.04$ & $27.63\pm3.51$ \\
%   \hline \hline
    \textbf{LERP} & $\mathbf{86.75\pm1.29}$ & $\mathbf{87.61\pm0.27}$ & $\mathbf{93.07\pm1.04}$ \\
    \hline
\end{tabular}}
\end{table}
%%%%%%%%%%%%%%%%%%%%%%%%%%%%%%%%%%%%%%%%%%%%%%%%%%%%%%%%%%%%%%

\section{Related Work}\label{sec:related_work}

%%%%%%%%%%%%%%%%%%%%%%%%%%%%%%%%%%%%%%%%%%%%%%%%%%%%%%%%%%%%%%%
%%%%%%%%%%%%%%%%%%%%%%%%%%%%%%%%%%%%%%%%%%%%%%%%%%%%%%%%%%%%%%%

\textbf{GCN-based Methods.} Graph convolutional networks (GCNs) have
achieved great success in semi-supervised graph learning
\cite{kipf2016semi, hamilton2017inductive, velickovic2018graph,
chen2018fastgcn}. Among them, the model proposed by Kipf \textit{et al.}
\cite{kipf2016semi}, which adopts the first-order approximation of
spectral graph convolution, offered good results in semi-supervised node
classification and revealed the identity between the spectral-based
graph convolution \cite{bruna2013spectral, defferrard2016convolutional}
and the spatial-based message-passing scheme \cite{gilmer2017neural,
scarselli2008graph}.  Afterward, there was an explosion of designs in
graph convolutional layers to offer various propagation mechanisms
\cite{monti2017geometric, hamilton2017inductive, velickovic2018graph,
xu2018powerful}. Numerous applications of graph-structured data arise
\cite{zhang2018end, xu2018powerful, you2018graphrnn}. 

However, graph regularization of label distributions in the design of
GCNs is not directly connected to label embeddings in model learning,
which could undermine the performance.  Some work attempts to address
this deficiency by proposing a giant graphic model that incorporates
label correlations \cite{qu2019gmnn, zhang2019bayesian, ma2019flexible}.
In practice, these methods are costly with a specific inductive bias and
subjective to the local minimum during optimization.  Another issue
induced is that a sufficient number of labels are required to learn the
feature transformation. To improve the label efficiency of GCNs, researchers
introduce other semi-supervised learning techniques \cite{li2018deeper,
sun2020multi} or improve layer-wise aggregation strength
\cite{li2019label, wan2021contrastive}.  However, they still inherit
deficiencies from graph convolutional layers. 

\textbf{Propagation-based Methods.} The propagation-based methods for
node classification on graphs can be dated back to the label propagation
algorithms in \cite{zhu2003semi, zhou2003learning}. Recently, there has
been a renaissance in combining propagation schemes with advanced neural
networks \cite{klicpera2018predict, huang2020combining,
klicpera2019diffusion, zhu2020simple, chen2020scalable}. These methods
attempt to encode higher-order correlations but preserve nodes' local
information simultaneously. For example, APPNP
\cite{klicpera2018predict} propagates neural predictions via label
propagation with small contributions of predictions added at each
iteration \cite{zhou2003learning}. It gives a superior performance as
compared to GCN-based methods. Yet, these methods do not target very
low label rates. Besides, the joint learning of feature transformation
still requires a considerable amount of labels for training. 

\section{Conclusion}\label{sec:conclusion}

A label efficient regularization and propagation framework was proposed
and a variational interpretation of GraphHop was given in this work.
Simply speaking, GraphHop can be viewed as an alternate optimization
process that optimizes a regularized function defined by graphs under
probability constraints.  Based on this understanding, two drawbacks of
GraphHop were identified, and a new method called LERP was proposed to
address them.  LERP solves two convex transformed subproblems in each
round with two salient features. First, it determines reliable nodes
adaptively for classifier training. Second, it adopts iterations to
efficiently offer an improved approximation to the optimal label
embeddings. The convergence was guaranteed by the mathematical proof,
and the performance of LERP was tested and compared with a large number
of existing methods against five commonly used datasets as well as an
object recognition problem. Its superior performance, especially at
extremely low label rates, demonstrates its effectiveness and
efficiency.  It would be interesting to apply other regularization
schemes to graph learning problems based on the framework of GraphHop
and LERP as a future extension. 

% or


\appendices
\section{Proof of Theorem \ref{theorem:convex_problem_w}}\label{app:convex_problem_w}
\begin{proof}
The cost function in Eq. (\ref{equ:graphhop_optimize_w_cross_entropy}) can be 
expressed as 
\begin{equation}
-\sum_{i=1}^nu_{\alpha, i}\sum_{j=1}^cf_{ij}\log(\sigma(\mathbf{z})_j),
\end{equation}
where $\mathbf{z} = \mathbf{W}\mathbf{f}_{M, i}^T$ and
$\sigma(\mathbf{z})_j = e^{z_j}\big/\sum_{k=1}^ce^{z_k}$ is the $j$th
entry of the vector applied the softmax function. Then, it can be
rewritten as
\begin{equation}
\sum_{i=1}^nu_{\alpha, i}\sum_{j=1}^cf_{ij}\left(\log\big(\sum_{k=1}^ce^{z_k}\big)
- z_j\right).
\end{equation}
It is proved in \cite{boyd2004convex} that the log-sum-exp function is a
convex function. A nonnegative linear combination of convex functions is
still a convex function. Thus, the cost function in Eq.
(\ref{equ:graphhop_optimize_w_cross_entropy}) is a convex function,
which leads to a convex optimization problem without any constraint. 
\end{proof}

\section{}
\subsection{Proof of Theorem \ref{theorem:convex_problem}}\label{app:convex_problem}
\begin{proof}
We first prove the cost function in Eq.
(\ref{equ:graphhop_optimize_f_fixed}) is a convex function. This is done by showing the second derivative of the cost function is positive semi-definite. Formally, we have 
\begin{equation}
\begin{aligned} &
\begin{aligned}
\nabla^2 & \bigg[ \text{tr}(\mathbf{F}^T\mathbf{\tilde{L}}\mathbf{F})
+ \text{tr}((\mathbf{F}-\mathbf{F}_{init})^T\mathbf{U}(\mathbf{F}-\mathbf{F}_{init})) \\
& + \text{tr}\left((\mathbf{F}-\sigma(\mathbf{F}_M\mathbf{W}^T))^T\mathbf{U}_\alpha(
\mathbf{F}-\sigma(\mathbf{F}_M\mathbf{W}^T))\right) \bigg]
\end{aligned} \\
& = \mathbf{\tilde{L}} + \mathbf{U} + \mathbf{U}_\alpha
\end{aligned}.
\end{equation}
The first term, $\mathbf{\tilde{L}}$, is the random-walk normalized Laplacian matrix with eigenvalues in the range $[0, 2]$. The second and the third terms are both diagonal matrix with positive entries. Therefore, their sum, i.e., the second derivative of the cost function, is positive semi-definite. Besides, since the equality constraint is a linear function and inequality constraint is convex, the domain of the cost function is also convex. So based on the second-order characterization, Eq. (\ref{equ:graphhop_optimize_f_fixed}) is a convex optimization problem.
\end{proof}

\subsection{Proof of Proposition \ref{proposition}}\label{app:proposition}
\begin{proof}
Without loss of generality, we set $\mathbf{F}^{(0)}=\mathbf{F}_{init}$. 
By Equation (\ref{equ:variational_lp}), we have
\begin{equation}
\mathbf{F}^{(t)} = (\mathbf{U}_\beta\mathbf{\tilde{A}})^{t-1}\mathbf{F}_{init} 
+ \sum_{i=0}^{t-1}(\mathbf{U}_\beta\mathbf{\tilde{A}})^i(\mathbf{I} 
- \mathbf{U}_\beta)\mathbf{Y}^\prime.
\end{equation}
Since each diagonal entry of $\mathbf{U}_\beta$ is in $(0, 1)$ and the
eigenvalues of $\mathbf{\tilde{A}}$ in $[-1, 1]$, we have
\begin{equation}
\lim_{t\rightarrow\infty}(\mathbf{U}_\beta\mathbf{\tilde{A}})^{t-1}=0\  
\text{and}\ \lim_{t\rightarrow\infty}\sum_{i=0}^{t-1}(\mathbf{U}_\beta 
\mathbf{\tilde{A}})^{i} = (\mathbf{I} - \mathbf{U}_\beta\mathbf{\tilde{A}})^{-1}.
\end{equation}
Hence, 
\begin{equation}
\begin{aligned}
\mathbf{F}^* & = (\mathbf{I}-\mathbf{U}_\beta\mathbf{\tilde{A}})^{-1}
(\mathbf{I}-\mathbf{U}_\beta)\mathbf{Y}^\prime \\
& = (\mathbf{I}-(\mathbf{I}+\mathbf{U}^\prime)^{-1}\mathbf{\tilde{A}})^{-1}
(\mathbf{I} - (\mathbf{I}+\mathbf{U}^\prime)^{-1})\mathbf{Y}^\prime \\
& = (\mathbf{U}^\prime + \mathbf{I} - \mathbf{\tilde{A}})^{-1}(\mathbf{I} 
+ \mathbf{U}^\prime)(\mathbf{I} + \mathbf{U}^\prime)^{-1}\mathbf{U}^\prime 
\mathbf{Y}^{\prime} \\
& = (\mathbf{U}^\prime+\mathbf{\tilde{L}})^{-1}\mathbf{U}^\prime\mathbf{Y}^\prime 
\end{aligned}
\end{equation}
\end{proof}

\subsection{Proof of Theorem \ref{theorem:iteration_optimal}}\label{app:iteration_optimal}
\begin{proof}
We first show by induction that the iteration in Eq.
(\ref{equ:variational_lp}) satisfies the probability constraints in Eq.
(\ref{equ:graphhop_optimize_f_fixed}). Based on the assumption, the
starting $\mathbf{F}^{(0)}$ satisfies the constraints. Note that
\begin{equation}
\resizebox{.99\hsize}{!}{$
\begin{aligned}
\mathbf{Y}^\prime\mathbf{1}_c & =(\mathbf{U} + \mathbf{U}_\alpha)^{-1}
\mathbf{U}\mathbf{F}_{init}\mathbf{1}_c +(\mathbf{U}+\mathbf{U}_\alpha)^{-1} 
\mathbf{U}_\alpha\sigma(\mathbf{F}_M\mathbf{W}^T)\mathbf{1}_c \\
& = (\mathbf{U} + \mathbf{U}_\alpha)^{-1}\mathbf{U}\mathbf{1}_n + (\mathbf{U} 
+ \mathbf{U}_\alpha)^{-1}\mathbf{U}_\alpha\mathbf{1}_n \\
& = \mathbf{1}_n
\end{aligned}$}.
\end{equation}
Now, we assume that variable $\mathbf{F}^{(t-1)}$ at iteration $(t-1)$
meets the probability constraints. At the next iteration $t$,  we have 
\begin{equation}
\begin{aligned}
\mathbf{F}^{(t)}\mathbf{1}_c & = \mathbf{U}_\beta\mathbf{\tilde{A}}\mathbf{F}^{(t-1)}
\mathbf{1}_c + (\mathbf{I} - \mathbf{U}_\beta)\mathbf{Y}^\prime\mathbf{1}_c \\
& = \mathbf{U}_\beta\mathbf{\tilde{A}}\mathbf{1}_n + (\mathbf{I}-\mathbf{U}_\beta)\mathbf{1}_n \\
& = \mathbf{U}_\beta\mathbf{1}_n + (\mathbf{I}-\mathbf{U}_\beta)\mathbf{1}_n \\
& = \mathbf{1}_n
\end{aligned}.
\end{equation}
Since all matrices have nonnegative entries with only sum and
multiplication operations, the constraint
\begin{equation}
\mathbf{F}^{(t)}\geq\mathbf{0}
\end{equation}
holds. By induction, the convergence result in Eq.
(\ref{equ:optimization_solution}) satisfies the probability constraints.
Based on the fact that the optimization problem in Eq.
(\ref{equ:graphhop_optimize_f_fixed}) is a convex optimization problem
and Proposition \ref{proposition}, the optimality that meets the
constraints of the cost function is the solution to Eq.
(\ref{equ:graphhop_optimize_f_fixed})
\end{proof}

\section{Proof of Theorem \ref{theorem:convergence}}\label{app:convergence}
\begin{proof}
Given the optimization variable $\mathbf{F}_r$ and $\mathbf{W}_r$ from
the $r$th alternate round, then for the $(r + 1)$th alternate round, we
have the corresponding variables $\mathbf{F}_{r + 1}$ and $\mathbf{W}_{r
+ 1}$, respectively. \\ 
When fixed $\mathbf{F}$, we update $\mathbf{W}$ and get
\begin{equation}\label{equ:w_inequality_convergence}
\begin{aligned} &
\begin{aligned}
\text{tr} & (\mathbf{F}_r^T\mathbf{\tilde{L}}\mathbf{F}_r)  
+ \text{tr}((\mathbf{F}_r-\mathbf{F}_{init})^T\mathbf{U}(\mathbf{F}_r-\mathbf{F}_{init}))\\
& + \text{tr}\left((\mathbf{F}_r-\sigma(\mathbf{F}_M\mathbf{W}_r^T))^T\mathbf{U}_\alpha(
\mathbf{F}_r-\sigma(\mathbf{F}_M\mathbf{W}_r^T))\right)
\end{aligned} \\
& 
\begin{aligned}
\geq \text{tr} & (\mathbf{F}_r^T\mathbf{\tilde{L}}\mathbf{F}_r) 
+ \text{tr}((\mathbf{F}_r-\mathbf{F}_{init})^T\mathbf{U}(\mathbf{F}_r-\mathbf{F}_{init}))\\
& + \text{tr}\left((\mathbf{F}_r-\sigma(\mathbf{F}_M\mathbf{W}_{r+1}^T))^T\mathbf{U}_\alpha(
\mathbf{F}_r-\sigma(\mathbf{F}_M\mathbf{W}_{r + 1}^T))\right).
\end{aligned}
\end{aligned}
\end{equation}
Obviously, Eq. (\ref{equ:w_inequality_convergence}) holds due to the
minimization of Eq. (\ref{equ:graphhop_plus_classifier_loss}). Note that
all constraints are satisfied and $\mathbf{F}_M = ||_{0\leq m\leq
M}\mathbf{\tilde{A}}^m\mathbf{F}_r$. Similarly, when fixed $\mathbf{W}$,
then updated $\mathbf{F}$, we have
\begin{equation}\label{equ:f_inequality_convergence}
\begin{aligned} &
\begin{aligned}
& \text{tr} (\mathbf{F}_r^T\mathbf{\tilde{L}}\mathbf{F}_r) 
+ \text{tr}((\mathbf{F}_r-\mathbf{F}_{init})^T\mathbf{U}(\mathbf{F}_r-\mathbf{F}_{init}))\\
& + \text{tr}\left((\mathbf{F}_r-\sigma(\mathbf{F}_M\mathbf{W}_{r+1}^T))^T\mathbf{U}_\alpha(
\mathbf{F}_r-\sigma(\mathbf{F}_M\mathbf{W}_{r + 1}^T))\right)
\end{aligned} \\
& 
\begin{aligned}
\geq & \text{tr}  (\mathbf{F}_{r + 1}^T\mathbf{\tilde{L}}\mathbf{F}_{r + 1}) 
+ \text{tr}((\mathbf{F}_{r + 1}-\mathbf{F}_{init})^T\mathbf{U}(\mathbf{F}_{r + 1}-\mathbf{F}_{init}))\\
& + \text{tr}\left((\mathbf{F}_{r + 1}-\sigma(\mathbf{F}_M\mathbf{W}_{r+1}^T))^T\mathbf{U}_\alpha(
\mathbf{F}_{r + 1}-\sigma(\mathbf{F}_M\mathbf{W}_{r + 1}^T))\right).
\end{aligned}
\end{aligned}
\end{equation}
Apparently, this inequality holds due to the iterative step in Eq.
(\ref{equ:graphhop_plus_iteration}), which results in a minimization to
problem (\ref{equ:graphhop_optimize_f_fixed}). Note that the iteration
in Eq. (\ref{equ:graphhop_plus_iteration}) follows the probability
constraints and $\mathbf{F}_M = ||_{0\leq m\leq
M}\mathbf{\tilde{A}}^m\mathbf{F}_r$.  By summing Eqs.
(\ref{equ:w_inequality_convergence}) and
(\ref{equ:f_inequality_convergence}), we have
\begin{equation} \label{equ:final_inequality_convergence}
\begin{aligned} &
\begin{aligned}
\text{tr} & (\mathbf{F}_r^T\mathbf{\tilde{L}}\mathbf{F}_r)  
+ \text{tr}((\mathbf{F}_r-\mathbf{F}_{init})^T\mathbf{U}(\mathbf{F}_r-\mathbf{F}_{init}))\\
& + \text{tr}\left((\mathbf{F}_r-\sigma(\mathbf{F}_M\mathbf{W}_r^T))^T\mathbf{U}_\alpha(
\mathbf{F}_r-\sigma(\mathbf{F}_M\mathbf{W}_r^T))\right)
\end{aligned} \\
& 
\begin{aligned}
\geq & \text{tr}  (\mathbf{F}_{r + 1}^T\mathbf{\tilde{L}}\mathbf{F}_{r + 1}) 
+ \text{tr}((\mathbf{F}_{r + 1}-\mathbf{F}_{init})^T\mathbf{U}(\mathbf{F}_{r + 1}-\mathbf{F}_{init}))\\
& + \text{tr}\left((\mathbf{F}_{r + 1}-\sigma(\mathbf{F}_M\mathbf{W}_{r+1}^T))^T\mathbf{U}_\alpha(
\mathbf{F}_{r + 1}-\sigma(\mathbf{F}_M\mathbf{W}_{r + 1}^T))\right).
\end{aligned}
\end{aligned}
\end{equation}
However, we observe that in the original problem
(\ref{equ:graphhop_variational}), $\mathbf{F}_M$ is calculated from the
current label embeddings $\mathbf{F}$. Appearing here should be the $(r
+ 1)$th alternate round $\mathbf{F}_{r + 1}$, where in Eq.
(\ref{equ:final_inequality_convergence}) it is computed from the past
$r$th alternate round $\mathbf{F}_r$. So we define the aggregation at
the $(r + 1)$th round as $\mathbf{F}_M^\prime = ||_{0\leq m\leq
M}\mathbf{\tilde{A}}^m\mathbf{F}_{r + 1}$. By taking the absolute
difference between the cost function in Eq.
(\ref{equ:graphhop_variational}) and the right hand side of Eq.
(\ref{equ:final_inequality_convergence}), we arrive at
\begin{equation} \label{equ:original_approx_difference}
\begin{aligned} &
\begin{aligned}
\bigg|\text{tr} & \left((\mathbf{F}_{r + 1}-\sigma(\mathbf{F}_M\mathbf{W}_{r+1}^T))^T\mathbf{U}_\alpha(
\mathbf{F}_{r + 1}-\sigma(\mathbf{F}_M\mathbf{W}_{r + 1}^T))\right) \\
& - \text{tr} \left((\mathbf{F}_{r + 1}-\sigma(\mathbf{F}_M^\prime\mathbf{W}_{r+1}^T))^T\mathbf{U}_\alpha(
\mathbf{F}_{r + 1}-\sigma(\mathbf{F}_M^\prime\mathbf{W}_{r + 1}^T))\right)\bigg|
\end{aligned} \\
& 
\begin{aligned}
= \bigg| & \sum_{i = 1}^n u_{\alpha, i}||\mathbf{f}_{r+1, i} - \sigma(\mathbf{W}_{r + 1}\mathbf{f}^T_{M, i})||^2_2 \\
& - \sum_{i = 1}^n u_{\alpha, i}||\mathbf{f}_{r + 1, i} - \sigma(\mathbf{W}_{r + 1}\mathbf{f}_{M, i}^{\prime^T})||^2_2\bigg|
\end{aligned} \\
&
\begin{aligned}
\leq \sum_{i = 1}^nu_{\alpha, i}||\sigma(\mathbf{W}_{r + 1}\mathbf{f}_{M, i}^{\prime^T}) - \sigma(\mathbf{W}_{r + 1}\mathbf{f}_{M, i}^T)||^2_2
\end{aligned} \\
&
\begin{aligned}
\leq \sum_{i = 1}^n 2u_{\alpha, i} = 2\text{tr}(\mathbf{U}_\alpha)
\end{aligned}
\end{aligned}
\end{equation}
The first inequality in Eq. (\ref{equ:original_approx_difference}) holds
because of the reverse triangle inequality, and the second inequality is
due to the fact that the maximum squared value of the $l_2$-norm between
the difference of two probability distributions is $2$. \\ Since problem
(\ref{equ:graphhop_variational}) has a lower bound $0$ in theory,
together considering the results in Eqs.
(\ref{equ:final_inequality_convergence}) and
(\ref{equ:original_approx_difference}), then LERP algorithm
\ref{alg:graphhop_plus} monotonically decreases the value of problem
(\ref{equ:graphhop_variational}) within a constant difference
$2\text{tr}(\mathbf{U}_\alpha)$ and with constraints satisfied in each
round until the algorithm converges. 
\end{proof}

\bibliographystyle{IEEEtran}
\bibliography{lerp}

% \end{IEEEbiography}


% % insert where needed to balance the two columns on the last page with
% %\newpage


\end{document}